\documentclass{article}

\usepackage{arxiv}

\usepackage[utf8]{inputenc} 
\usepackage[T1]{fontenc}    
\usepackage{hyperref}       
\usepackage{url}            
\usepackage{booktabs}       
\usepackage{amsfonts}       
\usepackage{nicefrac}       
\usepackage{microtype}      
\usepackage{graphicx}

\usepackage{mathtools,amssymb} 
\usepackage{textcomp,marvosym} 
\usepackage{bm, scalerel}
\DeclareMathOperator*{\argmin}{arg\,min}
\DeclareMathOperator*{\argmax}{arg\,max}
\DeclareMathOperator*{\E}{\mathbb{E}}

\usepackage[colorinlistoftodos]{todonotes}
\usepackage{algorithmic}
\usepackage{algorithm}

\usepackage{setspace}

\title{Deep hybrid models: infer and plan in a dynamic world}

\author{
	Matteo Priorelli \\
	Institute of Cognitive Sciences and Technologies\\
	National Research Council of Italy \\
	Sapienza University of Rome, Italy \\
	\texttt{matteo.priorelli@gmail.com} \\
	\And
	Ivilin Peev Stoianov \\
	Institute of Cognitive Sciences and Technologies\\
	National Research Council of Italy
	\texttt{ivilinpeev.stoianov@cnr.it} \\
}

\begin{document}
	
	\maketitle
	\setstretch{1.1}
	
	
	\begin{abstract}
		To determine an optimal plan for complex tasks, one often deals with dynamic and hierarchical relationships between several entities. Traditionally, such problems are tackled with optimal control, which relies on the optimization of cost functions; instead, a recent biologically-motivated proposal casts planning and control as an inference process. \textit{Active inference} assumes that action and perception are two complementary aspects of life whereby the role of the former is to fulfill the predictions inferred by the latter. Here, we present an active inference approach that exploits discrete and continuous processing, based on three features: the representation of \textit{potential body configurations} in relation to the objects of interest; the use of hierarchical relationships that enable the agent to easily interpret and flexibly expand its body schema for tool use; the definition of \textit{potential trajectories} related to the agent's intentions, used to infer and plan with dynamic elements at different temporal scales. We evaluate this \textit{deep hybrid model} on a habitual task: reaching a moving object after having picked a moving tool. We show that the model can tackle the presented task under different conditions. This study extends past work on planning as inference and advances an alternative direction to optimal control.
	\end{abstract}
	
	
	
	
	\section{Introduction}
	
	Imagine a baseball player striking a ball with a bat. State-of-the-art approaches to simulate such goal-directed movements in a human-like manner often rely on optimal control \cite{Todorov2004,diedrichsen2010coordination}. This theory rests upon the formulation of goals in terms of value functions and their optimization via cost functions. While this approach has advanced both robotics and our understanding of human motor control, its biological plausibility remains disputed \cite{Friston2012g}. Moreover, value functions for motor control might restrict the range of motions an agent can learn \cite{Friston2009a}. In contrast, complex movements like handwriting or walking can emerge naturally from generative models encoding goals as prior beliefs about the environment \cite{Friston2011opt}. The theory of active inference builds upon this premise, proposing that goal-directed behavior results from a biased internal representation of the world. This bias generates a cascade of prediction errors forcing the agent to sample those observations that make its beliefs true \cite{Friston2010,Friston2010a, Parr2021,Priorelli2023g}. In this view, the tradeoff between exploration and exploitation arises naturally, driving the agent to minimize the uncertainty of its internal model before maximizing potential rewards \cite{Parr2017}. Active inference shares foundational principles with predictive coding \cite{Rao1999,Clark2013} and the \textit{Bayesian brain hypothesis}, which postulates that the brain makes sense of the world by constructing a hierarchical generative model that continuously makes perceptual hypotheses and refines them \cite{Hohwy2013}. 
	
	This perspective offers a promising avenue for advancing current robotics and machine learning, particularly, in a research that frames control and planning as an inference process \cite{Millidge2020a,Botvinick2012,Toussaint2006,Toussaint2009,Stoianov2018}. A distinctive feature of active inference is its ability to model the environment as a hierarchy of causes and dynamic states evolving at different timescales \cite{Friston2008}, which is fundamental for biological phenomena such as linguistic communication \cite{Friston2020}, and for advanced movements such as that of a baseball player. In the latter, several characteristics of the nervous system can be well captured by an active inference agent, providing a robust alternative to optimal control. First, the human brain is assumed to maintain a hierarchical representation of the body that generates the motor commands required to achieve the desired goal \cite{Kandel2013}. This representation must be flexible, in that the baseball player should relate the configuration of the bat to his body, acting as an extension of his hand \cite{Cardinali2009}. Tool-use experiments in non-human primates revealed that parietal and motor regions rapidly adapt to incorporate tools into the body schema, enabling seamless interaction with objects \cite{Maravita2004}. Further, the player's brain should maintain a dynamic representation of the moving ball in order to predict its trajectory before and after the hit. The posterior parietal cortex is known to encode multiple objects in parallel during action sequences, forming visuomotor representations that also account for object affordances \cite{Baldauf2008}. In addition, distinct neural populations in the dorsal premotor cortex are known to encode multiple reaching options in decision-making tasks, with one population being activated and the others suppressed as the decision unfolds \cite{Cisek2005}. In short, the human brain can maintain multiple \textit{potential body configurations} and \textit{potential trajectories} appropriate for specific tasks.
	
	Despite its potential, the development of deep hierarchical models within active inference remains limited. Adaptation to complex data still relies primarily on neural networks as generative models \cite{Ueltzhoffer2017, Millidge2020, Fountas2020, Rood2020, Sancaktar2020, champion2023deconstructing, 9514227, ccatal2021robot, Yuan2023,Priorelli2023}. 
	One study demonstrated that a deep hierarchical agent, equipped with independent dynamics functions for each degree of freedom (DoF), could continuously adjust its internal trajectories to align with prior expectations, enabling advanced control of complex kinematic chains \cite{Priorelli2023b}. This ability to learn and act across intermediate timescales offers significant advantages for solving control tasks. 
	Moreover, simulating real-world scenarios can benefit from so-called \textit{hybrid} or \textit{mixed models} in active inference, which integrate discrete decision-making with continuous motion. However, state-of-the-art applications of such models have simulated static contexts only \cite{Parr2021,Friston2017,Friston2017a,Parr2018c,pf19}.
	
	In this work, we address these challenges from a unified perspective. Our key contributions are summarized as follows:
	\begin{itemize}
		\item We present an active inference agent affording robust planning in dynamic environments. The basic unit of this agent maintains potential trajectories, enabling a high-level discrete model to infer the state of the world and plan composite movements. The units are hierarchically combined to represent potential body configurations, incorporating object affordances (e.g., grasping a cup by the handle or with the whole hand) and their hierarchical relationships (e.g., how a tool can extend the agent's kinematic chain).
		
		\item We introduce a modular architecture designed for tasks that involve deep hierarchical modeling, such as tool use. Its multi-input and multi-output connectivity resembles traditional neural networks and represents an initial step toward designing deep structures in active inference capable of generalizing across and learning novel tasks.
		
		\item We evaluate the agent's performance in a common task: reaching a moving ball after reaching and picking a moving tool. The results highlight the interplay between the agent's potential trajectories and the dynamic accumulation of sensory evidence. We demonstrate the agent's ability to infer and plan under different conditions, such as random object positions and velocities.
	\end{itemize}
	
	
	\section{Methods}
	
	
	\subsection{\label{appendix:pcn}Predictive coding}
	
	According to predictive coding (PC), the human brain makes sense of the world by constructing an internal generative model of how hidden states of the environment generate the perceived sensations \cite{Rao1999,Clark2013,Hohwy2020}. This internal model is continuously refined by minimizing the discrepancy (called \textit{prediction error}) between sensations and the respective predictions. More formally, given some prior knowledge $p(\bm{x})$ over hidden states $\bm{x}$ and partial evidence $p(\bm{y})$ over sensations $\bm{y}$, the nervous system can find the posterior distribution of the hidden states given the sensations via Bayes rule:
	\begin{equation}
		p(\bm{x} | \bm{y}) = \frac{p(\bm{x}, \bm{y})}{p(\bm{y})}
	\end{equation}
	However, direct computation of the posterior $p(\bm{x} | \bm{y})$ is unfeasible since the evidence requires marginalizing over every possible outcome, i.e., $p(\bm{y})=\int{p(\bm{x}, \bm{y}) d\bm{x}}$. Predictive coding supposes that, instead of performing exact Bayesian inference, organisms are engaged in a variational approach \cite{Friston2009}, i.e., they approximate the posterior distribution with a simpler \textit{recognition} distribution $q(\bm{x}) \approx p(\bm{x} | \bm{y})$, and minimize the difference between the two distributions. This difference is expressed in terms of a Kullback-Leibler (KL) divergence:
	\begin{equation} 
		\label{eq:KL}
		D_{KL}[q(\bm{x})||p(\bm{x}|\bm{y})] = \int_{\bm{x}} q(\bm{x}) \ln \frac{q(\bm{x})}{p(\bm{x} | \bm{y})} d\bm{x}
	\end{equation}
	Given that the denominator $p(\bm{x}|\bm{y})$ still depends on the marginal $p(\bm{y})$, we express the KL divergence in terms of the log evidence and the \textit{Variational Free Energy} (VFE), and minimize the latter quantity instead. The VFE is the negative of what in the machine learning community is known as the \textit{evidence lower bound} or \textit{ELBO} \cite{Jordan1999}:
	\begin{equation}
		\mathcal{F} = \E_{q(\bm{x})} \left[ \ln \frac{q(\bm{x})}{p(\bm{x},\bm{y})} \right] = \E_{q(\bm{x})} \left[\ln \frac{q(\bm{x})}{p(\bm{x}|\bm{y})} \right] - \ln p(\bm{y})
	\end{equation}
	Since the KL divergence is always nonnegative, the VFE provides an upper bound on surprise, i.e., $\mathcal{F} \geq - \ln p(\bm{y})$. Therefore, minimizing $\mathcal{F}$ is equivalent to minimizing the KL divergence with respect to $q(\bm{x})$. The more the VFE approaches $0$, the closer the approximate distribution is to the real posterior, and the higher the model evidence (or, equivalently, the lower the surprise about sensory outcomes will be). The discrepancy between the two distributions also depends on the specific assumptions made over the recognition distribution: a common one is the Laplace approximation \cite{Friston2009}, which assumes Gaussian probability distributions, e.g., $q(\bm{x}) = \mathcal{N}(\bm{\mu}, \bm{\Sigma})$, where $\bm{\mu}$ represents the most plausible hypothesis - also called \textit{belief} about the hidden states $\bm{x}$ - and $\bm{\Sigma}$ is its covariance matrix. Now, we factorize the generative model $p(\bm{x}, \bm{y})$ into likelihood and prior terms, and we further parameterize it with some parameters $\bm{\theta}$:
	\begin{align}
		\begin{split}
			p(\bm{x}, \bm{y}) &= p(\bm{y} | \bm{x}, \bm{\theta}) p(\bm{x}) \\
			p(\bm{y} | \bm{x}, \bm{\theta}) &= \mathcal{N}(\bm{g}(\bm{x},\bm{\theta}), \bm{\Sigma}_y) \\
			p(\bm{x}) &= \mathcal{N}(\bm{\eta}, \bm{\Sigma}_\eta)
		\end{split}
	\end{align}
	where $\bm{g}(\bm{x},\bm{\theta})$ is a likelihood function and $\bm{\eta}$ is a prior. In this way, after applying the logarithm over the Gaussian terms, the VFE breaks down to the following simple formula:
	\begin{equation}
		\mathcal{F} = -\frac{1}{2} \left[ \bm{\Pi}_y \bm{\varepsilon}_y^2 + \bm{\Pi}_\eta \bm{\varepsilon}_\eta^2 + \ln 2 \pi \bm{\Sigma}_y + \ln 2 \pi \bm{\Sigma}_\eta \right]
	\end{equation}
	where we expressed the difference in the exponents of the Gaussian distributions in terms of \textit{prediction errors} $\bm{\varepsilon}_y = \bm{y} - \bm{g}(\bm{\mu},\bm{\theta})$ and $\bm{\varepsilon}_\eta = \bm{\mu} - \bm{\eta}$, and we wrote the covariances in terms of their inverse, i.e., precisions $\bm{\Pi}_y$ and $\bm{\Pi}_\eta$. In order to minimize the KL divergence between the real and approximate posteriors, we can minimize the VFE with respect to the beliefs $\bm{\mu}$ -- a process associated with \textit{perception} -- and the parameters $\bm{\theta}$ -- generally referred to as \textit{learning}. In practice, the update rules follow gradient descent:
	\begin{align}
		\begin{aligned}
			\bm{\mu} &= \argmin_{\bm{\mu}} \mathcal{F} \\
			\dot{\bm{\mu}} &= - \partial_{\mu} \mathcal{F} = \partial_{\mu} \bm{g}^T \bm{\Pi}_y \bm{\varepsilon}_y - \bm{\Pi}_\eta \bm{\varepsilon}_\eta
		\end{aligned}
		&&
		\begin{aligned}
			\bm{\theta} &= \argmin_{\bm{\theta}} \mathcal{F} \\
			\dot{\bm{\theta}} &= - \partial_{\theta} \mathcal{F} = - \partial_{\theta} \bm{g}^T \bm{\Pi}_y \bm{\varepsilon}_y
		\end{aligned}
	\end{align}
	In order to separate the timescales of fast perception and slow-varying learning, the two phases are treated as the steps of an EM algorithm, i.e., optimizing the beliefs while keeping the parameters fixed, and then optimizing the parameters while keeping the beliefs fixed.
	
	Predictive coding, originally rooted in data compression \cite{5391301}, has inspired biologically plausible alternatives to traditional deep learning, such as \textit{Predictive Coding Networks} (PCNs) \cite{Millidge2022a, salvatori2023braininspired}. In fact, the predictive coding algorithm can be scaled up to learn highly complex structures, composed of causal relationships of arbitrarily high depth -- in a similar way to deep neural networks. In particular, we can factorize a generative model into a product of different distributions, wherein a specific level only depends on the level above:
	\begin{align}
		\begin{split}
			p(\bm{x}^{(0)}, \dots, \bm{x}^{(l)}) &= p(\bm{x}^{(0)}) \prod_{l=1}^L p(\bm{x}^{(l)}|\bm{x}^{(l-1)}) \\
			p(\bm{x}^{(l)} |\bm{x}^{(l-1)}) &= \mathcal{N}(\bm{g}(\bm{x}^{(l-1)},\bm{\theta}^{(l-1)}), \bm{\Sigma}^{(l)})
		\end{split}
	\end{align}
	where $0$ indicates the highest level, and $L$ is the number of levels in the hierarchy. In this way, $\bm{x}^{(l)}$ acts as an observation for level $l-1$, and $\bm{g}(\bm{x}^{(l)},\bm{\theta}^{(l)})$ acts as a prior for level $l+1$. As a result, this factorization allows us to express the update of beliefs and parameters of a specific level only based on the level above and the level below -- with a simple VFE minimization as in the previous case. Differently from the backpropagation algorithm of neural networks, the message passing of prediction errors implements a biologically plausible Hebbian rule between presynaptic and postsynaptic activities. In fact, the predictions $\bm{u}$ computed by the likelihood functions are typically a weighted combination of neurons passed to a nonlinear activation function $\bm{\phi}$:
	\begin{equation}
		\bm{u}_{i}^{(l)} = \sum_{j=1}^{J} \bm{W}_{i,j}^{(l-1)} \bm{\phi} (\bm{x}_{j}^{(l-1)})
	\end{equation}
	where $W_{i,j}^{(l-1)}$ are the weights from neuron $j$ at level $l-1$ to neuron $i$ at level $l$. Crucially, what is backpropagated in PCNs are not signals detecting increasingly complex features, but messages representing how much the model is \textit{surprised} about sensory observations (e.g., a prediction equal to an observation means that the network structure is a good approximation of the real process, and no errors have to be conveyed).
	
	
	\subsection{\label{appendix:hai}Hierarchical active inference}
	
	The theory of active inference builds on the same assumptions of predictive coding, but with two critical differences. The first one -- which is actually shared with some implementations of predictive coding such as temporal PC \cite{Millidge2023} -- assumes that living organisms constantly deal with highly dynamic environments, and the internal models that they build must reflect the changes occurring in the real generative process \cite{Parr2018b}. This relation is usually expressed in terms of generalized coordinates of motion (encoding, e.g., position, velocity, acceleration, and so on) \cite{Friston2010gen}; consequently, the environment is modeled with the following nonlinear system:
	\begin{align}
		\begin{split}
			\label{eq:general}
			\tilde{\bm{y}} &= \tilde{\bm{g}}(\tilde{\bm{x}}) + \bm{w}_y \\
			\mathcal{D}\tilde{\bm{x}} &= \tilde{\bm{f}}(\tilde{\bm{x}}, \tilde{\bm{v}}) + \bm{w}_x
		\end{split}
	\end{align}
	where $\tilde{\bm{x}}$ are the generalized hidden states, $\tilde{\bm{v}}$ are the generalized hidden causes, $\tilde{\bm{y}}$ are the generalized sensory signals, $\mathcal{D}$ is a differential operator that shifts all the temporal orders by one, i.e.: $\mathcal{D}\tilde{\bm{x}} = [\bm{x}', \bm{x}'', \bm{x}''', \dots]$, and the letter $w$ indicates (Gaussian) noise terms. The likelihood function $\tilde{\bm{g}}$ defines how hidden states generate sensory observations (as in predictive coding), while the dynamics function $\tilde{\bm{f}}$ specifies the evolution of the hidden states (see \cite{parr2022active,Friston2022} for more details). The associated joint probability is factorized into independent distributions:
	\begin{equation}
		p(\tilde{\bm{y}}, \tilde{\bm{x}}, \tilde{\bm{v}}) = p(\tilde{\bm{y}} | \tilde{\bm{x}}) p(\tilde{\bm{x}} | \tilde{\bm{v}}) p(\tilde{\bm{v}})
	\end{equation}
	where each distribution is Gaussian:
	\begin{align}
		\begin{split}
			p(\tilde{\bm{y}} | \tilde{\bm{x}}) &= \mathcal{N}(\tilde{\bm{g}}(\tilde{\bm{x}}), \tilde{\bm{\Sigma}}_y) \\
			p(\mathcal{D}\tilde{\bm{x}} | \tilde{\bm{v}}) &= \mathcal{N}(\tilde{\bm{f}}(\tilde{\bm{x}}, \tilde{\bm{v}}), \tilde{\bm{\Sigma}}_x) \\
			p(\tilde{\bm{v}} | \bm{\eta}) &= \mathcal{N}(\bm{\eta}, \tilde{\bm{\Sigma}}_v)
		\end{split}
	\end{align}
	As in predictive coding, these distributions are inferred through approximate posteriors $q(\tilde{\bm{x}})$ and $q(\tilde{\bm{v}})$, minimizing the related VFE $\mathcal{F}$.  As a result, the updates of the beliefs $\tilde{\bm{\mu}}$ and $\tilde{\bm{\nu}}$ respectively over the hidden states and hidden causes become:
	\begin{align}
		\begin{split}
			\dot{\tilde{\bm{\mu}}} - \mathcal{D} \tilde{\bm{\mu}} &= - \partial_\mu \mathcal{F} = \partial \tilde{\bm{g}}^T \tilde{\bm{\Pi}}_y \tilde{\bm{\varepsilon}}_y + \partial_\mu \tilde{\bm{f}}^T \tilde{\bm{\Pi}}_x \tilde{\bm{\varepsilon}}_x - \mathcal{D}^T \tilde{\bm{\Pi}}_x \tilde{\bm{\varepsilon}}_x \\
			\dot{\tilde{\bm{\nu}}} - \mathcal{D} \tilde{\bm{\nu}} &= - \partial_\nu \mathcal{F} = \partial_\nu \tilde{\bm{f}}^T \tilde{\bm{\Pi}}_x \tilde{\bm{\varepsilon}}_x - \tilde{\bm{\Pi}}_v \tilde{\bm{\varepsilon}}_v
		\end{split}
	\end{align}
	where $\tilde{\bm{\Pi}}_y$, $\tilde{\bm{\Pi}}_x$, and $\tilde{\bm{\Pi}}_v$ are the precisions, and $\tilde{\bm{\varepsilon}}_y$, $\tilde{\bm{\varepsilon}}_x$, and $\tilde{\bm{\varepsilon}}_v$ are respectively the prediction errors of sensory signals, dynamics, and priors:
	\begin{equation}
		\begin{split}
			\tilde{\bm{\varepsilon}}_y &= \tilde{\bm{y}} - \tilde{\bm{g}}(\tilde{\bm{\mu}}) \\
			\tilde{\bm{\varepsilon}}_x &= \mathcal{D} \tilde{\bm{\mu}} - \tilde{\bm{f}}(\tilde{\bm{\mu}}, \tilde{\bm{\nu}}) \\
			\tilde{\bm{\varepsilon}}_v &= \tilde{\bm{\nu}} - \bm{\eta}
		\end{split}
	\end{equation}
	For a full account of free energy minimization in active inference, see \cite{Parr2021}. Unlike the update rules of predictive coding, additional terms arise from the inferred dynamics, through which the model can capture the evolution of the environment. Also note that since we are minimizing over (dynamic) paths and not (static) states, an additional term $\mathcal{D} \tilde{\bm{\mu}}$ is present: this implies trajectory tracking, during which the VFE is minimized only when the belief of the generalized hidden states $\mathcal{D} \tilde{\bm{\mu}}$ matches its instantaneous trajectory $\dot{\tilde{\bm{\mu}}}$.
	
	The second assumption made by active inference is that our brains not only perceive (and learn) the external generative process, but also interact with it to reach desired states (e.g., in order to survive, not only one must understand which cause leads to an increase or decrease in temperature, but also take actions to live in a narrow range around 37 degrees Celsius). The \textit{free energy principle}, which is at the core of active inference, states that all living organisms act to minimize the free energy (or, equivalently, surprise). In fact, in addition to the perceptual inference of predictive coding, the VFE can be minimized by sampling those sensory observations that conform to some prior beliefs, e.g., $\bm{a} = \argmin_{\bm{a}} \mathcal{F}$, where $\bm{a}$ are the motor commands. This process is typically called \textit{self-evidencing}, implying that if I believe to find myself in a narrow range of temperatures, minimizing surprise via action will lead to finding places with such temperatures -- hence making my beliefs true. One role of the hidden causes is to define the agent's priors that ensure survival. These priors generate proprioceptive predictions which are suppressed by motor neurons via classical reflex arcs \cite{Adams2013}:	
	\begin{equation}
		\dot{\bm{a}} = - \partial_a \mathcal{F}_p = - \partial_a \tilde{\bm{y}}_p \tilde{\bm{\Pi}}_p \tilde{\bm{\varepsilon}}_p
	\end{equation}
	where $\partial_a \tilde{\bm{y}}_p$ is an inverse model from (proprioceptive) observations to actions, and $\tilde{\bm{\varepsilon}}_p=\tilde{\bm{y}}_p - \tilde{\bm{g}}_p(\tilde{\bm{\mu}})$ are the generalized proprioceptive prediction errors.
	
	As in predictive coding, we can scale up this generative model to capture the causal relationships of several related entities (e.g., the joints of a kinematic structure), up to a certain depth \cite{Friston2017, Friston2008, Pezzulo2018,Priorelli2023b}. Therefore, the prior becomes the prediction from the layer above, while the observation becomes the likelihood of the layer below:
	\begin{align}
		\begin{split}
			\dot{\tilde{\bm{\mu}}}^{(l)} &= \mathcal{D} \tilde{\bm{\mu}}^{(l)} + \partial \tilde{\bm{g}}^{(l)T} \tilde{\bm{\Pi}}_v^{(l)} \tilde{\bm{\varepsilon}}_v^{(l)} + \partial_\mu \tilde{\bm{f}}^{(l)T} \tilde{\bm{\Pi}}_x^{(l)} \tilde{\bm{\varepsilon}}_x^{(l)} - \mathcal{D}^T \tilde{\bm{\Pi}}_x^{(l)} \tilde{\bm{\varepsilon}}_x^{(l)} \\
			\dot{\tilde{\bm{\nu}}}^{(l)} &= \mathcal{D} \tilde{\bm{\nu}}^{(l)} + \partial_\nu \tilde{\bm{f}}^{(l)T} \tilde{\bm{\Pi}}_x^{(l)} \tilde{\bm{\varepsilon}}_x^{(l)} - \tilde{\bm{\Pi}}_v^{(l-1)} \tilde{\bm{\varepsilon}}_v^{(l-1)}
		\end{split}
	\end{align}
	where:
	\begin{equation}
		\begin{split}
			\tilde{\bm{\varepsilon}}_x^{(l)} &= \mathcal{D} \tilde{\bm{\mu}}_x^{(l)} - \tilde{\bm{f}}^{(l)}(\tilde{\bm{\mu}}^{(l)}, \tilde{\bm{\nu}}^{(l)}) \\
			\tilde{\bm{\varepsilon}}_v^{(l)} &= \tilde{\bm{\mu}}_v^{(l+1)} - \tilde{\bm{g}}^{(l)}(\tilde{\bm{\mu}}^{(l)})
		\end{split}
	\end{equation}
	and the subscript indicates the index of the level, as before. Here, the role of the hidden causes is to link hierarchical levels. This allows the agent to construct a hierarchy of representations varying at different temporal scales, critical for realizing richly structured behaviors such as linguistic communication \cite{Friston2020} or singing \cite{Friston2015}. In this case, the motor commands minimize the generalized prediction errors of the lowest level of the hierarchy. \\
	
	This continuous formulation of active inference is highly effective for dealing with realistic environments. However, minimizing the VFE (which is the free energy of the past and present) does not afford planning and decision-making in the immediate future. To do this, we construct a discrete generative model in which we discretize the possible future states and encode their expectations with categorical distributions. We then minimize the \textit{Expected Free Energy} (EFE) which, as the name suggests, is the free energy that the agents expect to perceive in the future \cite{Parr2019, Parr2021}. This discrete generative model is similar to the continuous model defined above, with the difference that we condition the hidden states over policies $\bm{\pi}$ (which in active inference are sequences of discrete actions):
	\begin{equation}
		p(\bm{s},\bm{o},\bm{\pi}) = p(\bm{o}|\bm{s},\bm{\pi}) p(\bm{s}|\bm{\pi}) p(\bm{\pi})
	\end{equation}
	Here, $\bm{s}$ and $\bm{o}$ are discrete states and outcomes, which represent past, present, and future states as in Hidden Markov Models. Then, the EFE is specifically constructed by considering future states as random variables that need to be inferred: 
	\begin{equation}
		\mathcal{G}_\pi = \E_{q(\bm{s},\bm{o}|\bm{\pi})} \left[ \ln \frac{q(\bm{s}|\bm{\pi})}{p(\bm{s},\bm{o}|\bm{\pi})} \right] \approx \E_{q(\bm{s},\bm{o}|\bm{\pi})} \left[ \ln \frac{q(\bm{s})}{q(\bm{s}|\bm{o},\bm{\pi})} \right] - \E_{q(\bm{o}|\bm{\pi})} [\ln p(\bm{o}|\bm{C})]    
	\end{equation}
	where $q(\bm{o}|\bm{\pi})$ is a recognition distribution that the agent constructs to infer the real posterior of the generative process.	Critically, the probability distribution $p(\bm{o}|\bm{C})$ encodes preferred outcomes, acting similar to a prior over the hidden causes in the continuous counterpart. The last two terms are respectively called \textit{epistemic} (uncertainty reducing) and \textit{pragmatic} (goal seeking). In practice, this quantity is used by first factorizing the agent's generative model as in POMDPs:
	\begin{equation}
		p(\bm{s}_{1:T},\bm{o}_{1:T},\bm{\pi}) = p(\bm{s}_1) \cdot p(\bm{\pi}) \cdot \prod_{\tau=1}^T p(\bm{o}_\tau|\bm{s}_\tau) \cdot \prod_{\tau=2}^T p(\bm{s}_\tau|\bm{s}_{\tau-1},\bm{\pi})
	\end{equation}
	Each of these elements can be represented with categorical distributions:
	\begin{align}
		\begin{aligned}
			p(\bm{s}_1) &= Cat(\bm{D}) \\
			p(\bm{\pi}) &= Cat(\bm{E})
		\end{aligned}
		&&
		\begin{aligned}
			p(\bm{o}_\tau|\bm{s}_\tau) &= Cat(\bm{A}) \\
			p(\bm{s}_\tau|\bm{s_}{\tau-1},\bm{\pi}) &= Cat(\bm{B}_{\pi,\tau})
		\end{aligned}
	\end{align}
	where $\bm{D}$ encodes beliefs about the initial state, $\bm{E}$ encodes the prior over policies, $\bm{A}$ is the likelihood matrix and $\bm{B}_{\pi,\tau}$ is the transition matrix. The minimization of EFE in discrete models follows the variational method used by predictive coding and continuous-time active inference; specifically, the use of categorical distributions breaks down the inference of hidden states and policies to a simple local message passing:
	\begin{align}
		\label{eq:disc_update}
		\begin{split}
			\bm{s}_{\pi,\tau} &= \sigma(\ln \bm{B}_{\pi,\tau-1} \bm{s}_{\pi,\tau-1} + \ln \bm{B}^T_{\pi,\tau} \bm{s}_{\pi,\tau+1} + \ln \bm{A}^T \bm{o}_\tau) \\
			\bm{\pi} &= \sigma(\ln \bm{E} - \mathcal{G}) \\
			\mathcal{G}_{\pi} &\approx \sum_\tau \bm{A} \bm{s}_{\pi,\tau} (\ln \bm{A} \bm{s}_{\pi,\tau} - \ln p(\bm{o}_\tau|\bm{C})) - \bm{s}_{\pi,\tau} diag(\bm{A}^T \ln \bm{A})
		\end{split}
	\end{align}
	For a complete treatment of how the EFE and the approximate posteriors are computed -- along with other insightful implications of the free energy principle in discrete models -- see \cite{parr2022active}, while \cite{Smith2022} provides a more practical tutorial with basic applications. Equation \ref{eq:disc_update} shows that the update rule for the discrete hidden states at time $\tau$ and conditioned over a policy $\pi$ is a combination of messages coming from the previous and next discrete time steps, and a message coming from the discrete outcome. This combination is passed to a softmax function in order to get a proper probability. In addition, the optimal policy $\bm{\pi}$ is found by a combination of a policy prior and the EFE, where the latter is a composition of epistemic and pragmatic behaviors. Policy inference can be refined by computing the EFE of several steps ahead in the future -- a process called \textit{sophisticated inference}. Then, at each discrete step $\tau$, the agents selects the most likely action $u$ under all policies, i.e., $u_t = \argmax_u \bm{\pi} \cdot [U_{\pi,t} = u]$.
	
	As before, we can construct a hierarchical structure able to express more and more invariant representations of hidden states \cite{DaCosta2020}. In discrete state-space, links between hierarchical levels are usually done between hidden states via the matrix $\bm{D}$ -- although in some formulations the hidden states of a level condition the policies of the subordinate levels. The update rules for this hierarchical alternative become:
	\begin{align}
		\begin{split}
			\bm{s}_{\pi,\tau}^{(l)} &= \sigma(\ln \bm{B}_{\pi,\tau-1}^{(l)} \bm{s}_{\pi,\tau-1}^{(l)} + \ln \bm{B}_{\pi,\tau}^{(l)T} \bm{s}_{\pi,\tau+1}^{(l)} + \ln \bm{A}^{(l)T} \bm{o}_\tau^{(l)} + \ln \bm{D}^{(l+1)T} \bm{s}_1^{(l+1)}) \\
			\bm{\pi}^{(l)} &= \sigma(\ln \bm{E}^{(l)} - \mathcal{G}^{(l)}) \\
			\mathcal{G}_{\pi}^{(l)} &\approx \sum_\tau \bm{A}^{(l)} \bm{s}_{\pi,\tau}^{(l)} (\ln \bm{A}^{(l)} \bm{s}_{\pi,\tau}^{(l)} - \ln p(\bm{o}^{(l)}_\tau|\bm{C})) - \bm{s}^{(l)}_{\pi,\tau} diag(\bm{A}^{(l)T} \ln \bm{A}^{(l)})
		\end{split}
	\end{align}
	This implies that each level takes abstract actions to minimize its EFE, only depending on the levels immediately below and above.
	
	
	\subsection{\label{appendix:bmc}Bayesian model comparison}
	
	Bayesian model comparison is a technique used to compare a posterior over some data with a few simple hypotheses known a-priori \cite{Friston2018a}. Consider a generative model $p(\bm{y}, \bm{\theta})$ with parameters $\bm{\theta}$ and data $\bm{y}$:
	\begin{equation}
		p(\bm{\theta},\bm{y}) = p(\bm{y}|\bm{\theta}) p(\bm{\theta})
	\end{equation}
	We introduce additional distributions $p(\bm{y},\bm{\theta} | m)$ which are reduced versions of the first model if the likelihood of some data is the same under both models -- i.e., $p(\bm{y}|\bm{\theta},m)=p(\bm{y}|\bm{\theta})$ -- and the only difference rests upon the specification of the priors $p(\bm{\theta} | m)$. We can express the posterior of the reduced models in terms of the posterior of the full model and the ratios of the priors and the evidence:
	\begin{equation}
		p(\bm{\theta}|\bm{y}, m) = p(\bm{\theta}|\bm{y}) \frac{p(\bm{\theta} | m) p(\bm{y})}{p(\bm{\theta}) p(\bm{y} | m)}
	\end{equation}
	The procedure for computing the reduced posteriors is the following: first, we integrate over the parameters to obtain the evidence ratio of the two models:
	\begin{equation}
		p(\bm{y} | m) = p(\bm{y}) \int p(\bm{\theta}|\bm{y}) \frac{p(\bm{\theta} | m)}{p(\bm{\theta})} d\bm{\theta}
	\end{equation}
	Then, we define an approximate posterior $q(\bm{\theta})$ and we compute the reduced free energies of each model $m$:
	\begin{equation}
		\label{eq:fe}
		\mathcal{F}[p(\bm{\theta} | m)] \approx \mathcal{F}[p(\bm{\theta})] + \ln \E_q \left[ \frac{p(\bm{\theta} | m)}{p(\bm{\theta})} \right]
	\end{equation}
	This VFE acts as a hint to how well the reduced representation explains the full model. Similarly, the approximate posterior of the reduced model can be written in terms of the posterior of the full model:
	\begin{equation}
		\label{eq:post}
		\ln q(\bm{\theta} | m) = \ln q(\bm{\theta}) + \ln \frac{p(\bm{\theta} | m)}{p(\bm{\theta})} - \ln \E_q \left[ \frac{p(\bm{\theta} | m)}{p(\bm{\theta})} \right]
	\end{equation}
	The Laplace approximation \cite{Friston2007} leads to a simple form of the approximate posterior and the reduced free energy. Assuming the following Gaussian distributions:
	\begin{align}
		\begin{aligned}
			p(\bm{\theta}) &= \mathcal{N}(\bm{\eta}, \bm{\Sigma}) \\
			p(\bm{\theta} | m) &= \mathcal{N}(\bm{\eta}_m, \bm{\Sigma}_m)
		\end{aligned}
		&&
		\begin{aligned}
			q(\bm{\theta}) &= \mathcal{N}(\bm{\mu}, \bm{C}) \\
			q(\bm{\theta} | m) &= \mathcal{N}(\bm{\mu}_m, \bm{C}_m)
		\end{aligned}
	\end{align}
	the reduced free energy turns to:
	\begin{align}
		\begin{split}
			\mathcal{F}[p(\bm{\theta}|m)] & \approx \mathcal{F}[p(\bm{\theta})] + \frac{1}{2} \ln |\bm{\Pi}_m \bm{P} \bm{C}_m \bm{\Sigma}| \\ 
			& -\frac{1}{2} (\bm{\mu}^T \bm{P} \bm{\mu} - \bm{\mu}_m^T \bm{P}_m \bm{\mu}_m - \bm{\eta}^T \bm{\Pi} \bm{\eta} + \bm{\eta}_m^T \bm{\Pi}_m \bm{\eta}_m)
		\end{split}
	\end{align}
	expressed in terms of precision of priors $\bm{\Pi}$ and posteriors $\bm{P}$. The reduced posterior mean and precision are computed via Equation \ref{eq:post}:
	\begin{align}
		\begin{split}
			\label{eq:mu_m}
			\bm{\mu}_m &= \bm{C}_m (\bm{P} \bm{\mu} - \bm{\Pi} \bm{\eta} + \bm{\Pi}_m \bm{\eta}_m) \\
			\bm{P}_m &= \bm{P} - \bm{\Pi} + \bm{\Pi}_m
		\end{split}
	\end{align}
	In this way, two different hypotheses $i$ and $j$ can be easily compared to infer which is the most likely to have generated the observed data; this comparison has the form of a log-Bayes factor $\mathcal{F}[p(\bm{\theta}|i)]- \mathcal{F}[p(\bm{\theta}|j]$. For a more detailed treatment of Bayesian model comparison (in particular under the Laplace approximation), see \cite{Friston2018a,Friston2011c}.
	
	
	
	\section{Results}
	
	\subsection{Deep hybrid models}
	
	Analyzing the emergence of distributed intelligence, Friston et al. \cite{FRISTON2024105500} emphasized three kinds of depth within the framework of active inference: factorial, hierarchical, and temporal. Factorial depth assumes independent factors in the agent's generative model (e.g., objects and qualities of an environment, or more abstract states), which can be combined to generate outcomes and transitions. Hierarchical depth introduces causal relationships between levels, inducing a separation of temporal scales whereby higher levels happen to construct more invariant representations, while lower levels better capture the rapid changes of sensory stimuli. Temporal depth entails, in discrete terms, a vision into the imminent future that can be used for decision-making; or, in continuous terms, increasingly precise estimates of dynamic trajectories.
	
	In the following, we present the main features of a \textit{deep hybrid model} in terms of factorial, temporal, and hierarchical depths in the context of flexible behavior, iterative transformations of reference frames, and dynamic planning. By deep hybrid model, we intend an active inference model composed of hybrid units connected hierarchically. Here, hybrid means that discrete and continuous representations are encoded within each unit, wherein the communication between the two domains is achieved by Bayesian model reduction \cite{Friston2011c,Friston2018a}. As a technical note, all the internal operations can be computed through automatic differentiation, i.e., by maintaining the gradient graph when performing each forward pass and propagating back the prediction errors. For a detailed treatment of predictive coding, hierarchical active inference in discrete and continuous state-spaces, and Bayesian model comparison, see Sections \ref{appendix:pcn}, \ref{appendix:hai}, and \ref{appendix:bmc}, respectively.
	
	
	\subsubsection{\label{section:flexible}Factorial depth and flexible behavior}
	
	\begin{figure}
		\centering
		\begin{minipage}[b]{.45\textwidth}
			\centering
			{\label{fig:hybrid_graph}\includegraphics[width=\textwidth]{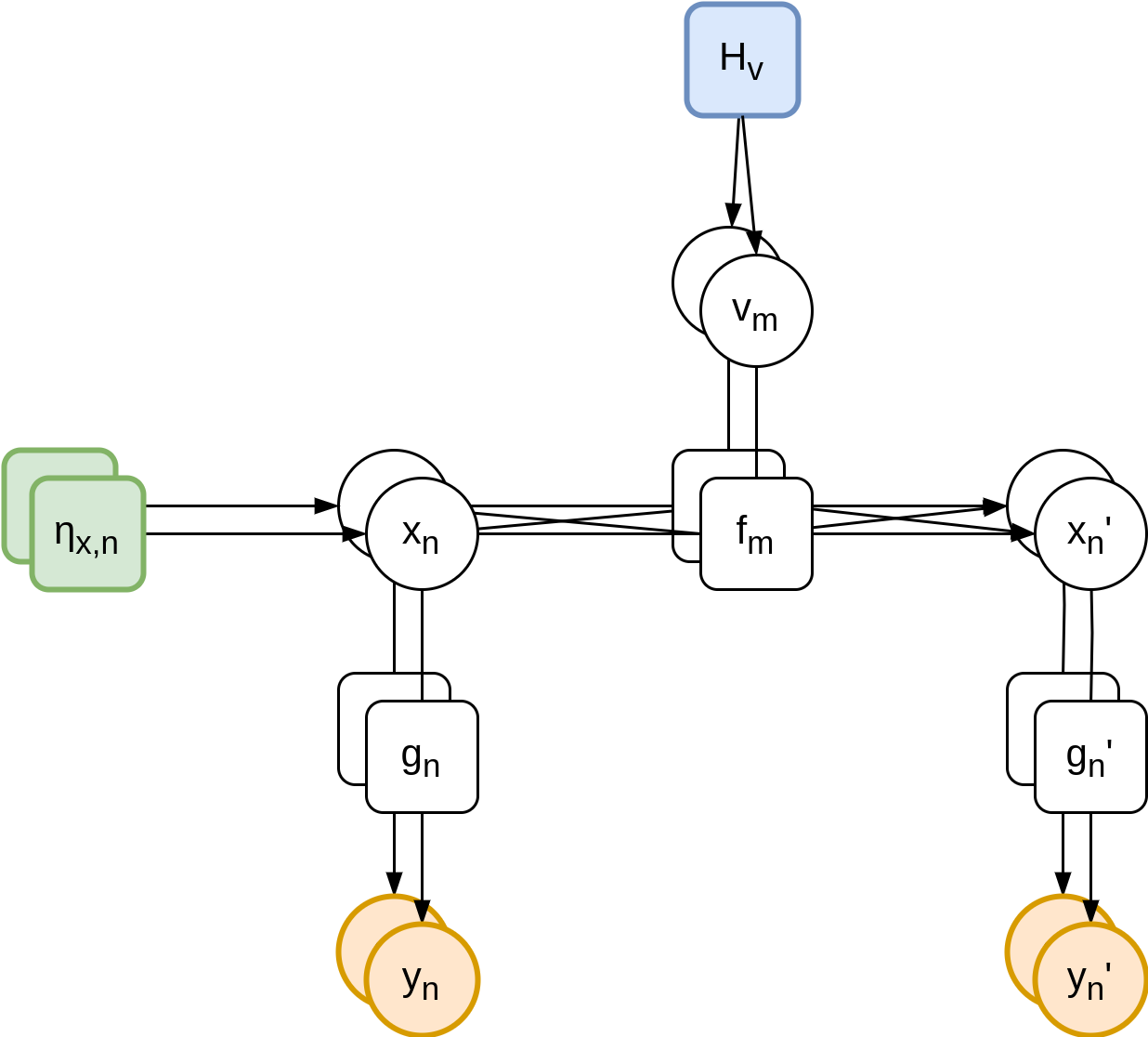}}
		\end{minipage}
		\hfill
		\begin{minipage}[b]{.45\textwidth}
			{\label{fig:frames_dyn_inf}\includegraphics[width=\textwidth]{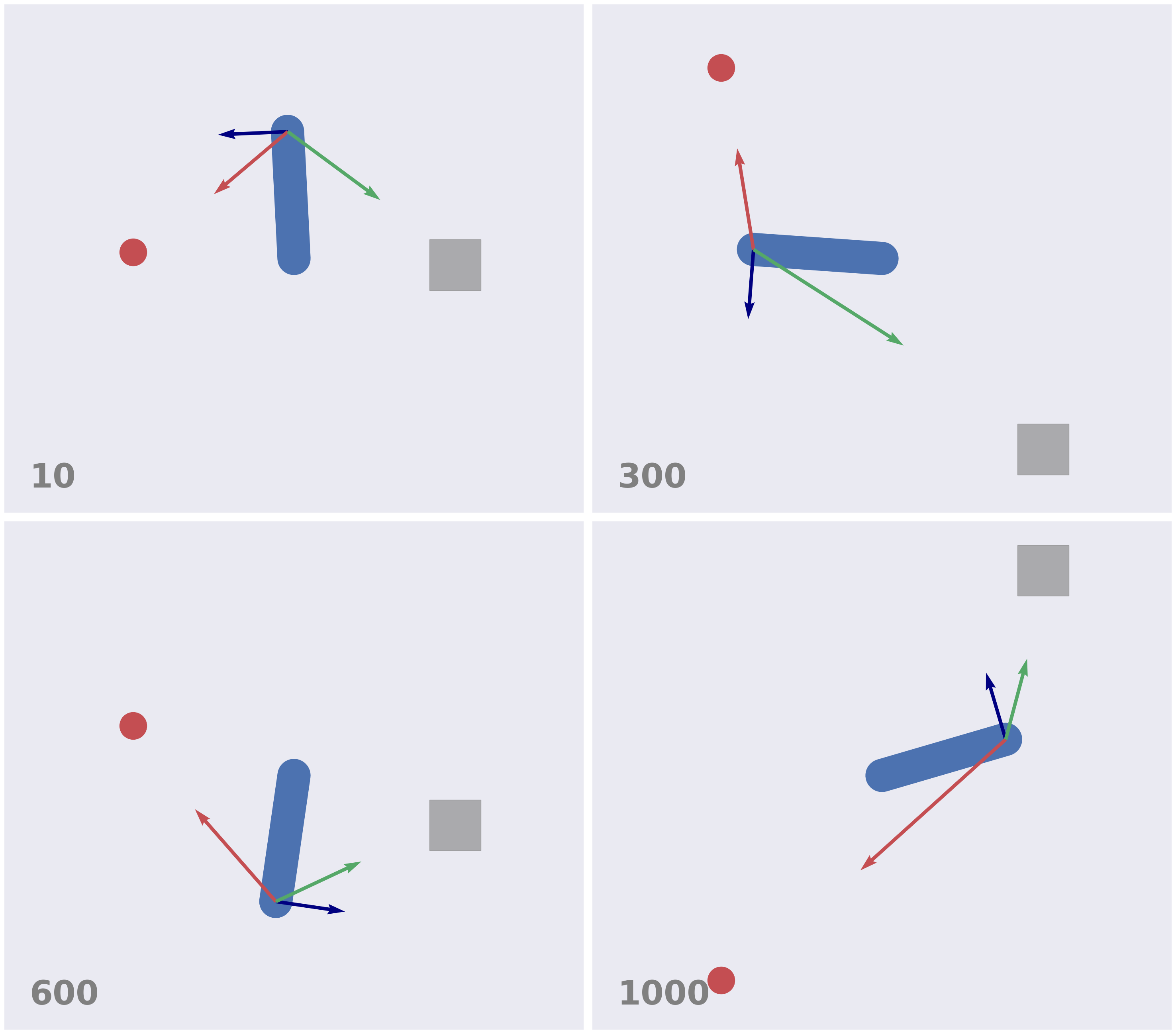}}
		\end{minipage}
		\caption{\textbf{(a)} Factor graph of a hybrid unit. Continuous hidden states $\tilde{\bm{x}}$ generate predictions $\tilde{\bm{y}}$ through parallel pathways. Model dynamics is encoded by potential trajectories $\bm{f}_m$, which are hypotheses of how the world may evolve and are associated with discrete hidden causes $\bm{v}$. \textbf{(b)} Illustrative example of a hybrid unit. In this task, the agent has to infer which one among two objects (a red circle and a gray square moving along a circular trajectory) is being tracked by another 1-DoF agent. The time step is shown in the bottom left of each frame. The hidden states $\tilde{\bm{x}}$ encode the angle and angular velocity of the arm (generating proprioceptive predictions), as well as the positions and velocities of the two objects (generating visual predictions). The blue arrow represents the actual hand trajectory, while the red and green arrows represent the two potential trajectories associated with reaching movements toward the two objects. See \cite{Priorelli2023e} for more details.}
		\label{fig:hybrid}
	\end{figure}
	
	Consider the case where one needs to infer which object is being followed by an agent. This can be done through a \textit{hybrid unit} $\mathcal{U}$, whose factor graph is depicted in Figure \ref{fig:hybrid_graph}. The variables are: continuous hidden states $\tilde{\bm{x}}=[\bm{x}, \bm{x}^\prime]$, observations $\tilde{\bm{y}}=[\bm{y}, \bm{y}^\prime]$, and (discrete) hidden causes $\bm{v}$. Hidden states and observations comprise two temporal orders (e.g., $\bm{x}$ encodes the position and $\bm{x}^\prime$ the velocity), and the first temporal order will be indicated as the 0th order -- see the Methods section for more information. The factors are: dynamics functions $\bm{f}$, and likelihood functions $\tilde{\bm{g}}=[\bm{g}, \bm{g}^\prime]$. Note also a prior over the 0th-order hidden states $\bm{\eta}_x$, and a prior over the hidden causes $\bm{H}_v$ encoding the agent's goals. The generative model is the following:
	\begin{equation}
		p(\tilde{\bm{x}}, \bm{v}, \tilde{\bm{y}}) = p(\tilde{\bm{y}}|\tilde{\bm{x}}) p(\bm{x}^\prime|\bm{x}, \bm{v}) p(\bm{x}) p(\bm{v})
	\end{equation}
	We assume that hidden causes and hidden states have different dimensions, resulting in two factorizations. The hidden states, with dimension $N$, are sampled from independent Gaussian distributions and generate predictions in parallel pathways:
	\begin{align}
		p(\bm{x}) &= \prod_n^N \mathcal{N}(\bm{\eta}_{x,n}, \bm{\Sigma}_{\eta,x,n}) &
		p(\tilde{\bm{y}}|\tilde{\bm{x}}) &= \prod_n^N \mathcal{N}(\tilde{\bm{g}}_n(\tilde{\bm{x}}_n), \tilde{\bm{\Sigma}}_{y,n})
	\end{align}
	where $\bm{\Sigma}_{\eta,x,n}$ and $\tilde{\bm{\Sigma}}_{y,n}$ are their covariance matrices. In turn, the hidden causes, with dimension $M$, are sampled from a categorical distribution:
	\begin{equation}
		p(\bm{v}) = Cat(\bm{H}_v)
	\end{equation}
	This differs from state-of-the-art hybrid architectures which assume separate continuous and discrete models with continuous hidden causes (see Section \ref{appendix:hai}) \cite{Parr2018b}. A discrete hidden cause $v_m$ concurs, with hidden states $\bm{x}$, in generating a specific prediction for the 1st temporal order $\bm{x}^\prime$:
	\begin{equation}
		p(\bm{x}^\prime|\bm{x}, m) = \mathcal{N}(\bm{f}_m(\bm{x}), \bm{\Sigma}_{x,m})
	\end{equation}
	This probability distribution entails a \textit{potential trajectory}, or hypothetical evolution of the hidden states, which the agent maintains to infer the state of affairs of the world and act. More formally, we consider $p(\bm{x}^\prime|\bm{x}, m)$ as being the $m$th reduced version of a full model:
	\begin{equation}
		p(\bm{x}' | \bm{x}, \bm{v}) = \mathcal{N}(\bm{\eta}_{x}^\prime, \bm{\Sigma}_x)
	\end{equation}
	This allows us -- using the variational approach for approximating the true posterior distributions -- to convert discrete signals into continuous signals and vice versa through Bayesian model average and Bayesian model comparison, respectively (see Section \ref{appendix:bmc} and \cite{Friston2011c,Friston2018a}). In particular, top-down messages combine the potential trajectories $\bm{f}_m(\bm{x})$ with the related probabilities encoded in the discrete hidden causes $\bm{v}$:
	\begin{equation}
		\bm{\eta}_{x}^\prime = \sum_m^M v_m \bm{f}_m(\bm{x})
	\end{equation}
	This computes a dynamic path that is an average of the agent's hypotheses, based on its prior $\bm{H}_v$. Conversely, bottom-up messages compare the agent's prior surprise $-\ln \bm{H}_v$ with the log evidence $l_m$ of every reduced model, i.e., $\bm{v} = \sigma(\ln \bm{H}_v + \bm{l})$, where $\bm{l} = [l_1, \dots, l_M]$ and $\sigma$ is a softmax function. The log evidence is accumulated over a continuous time $T$:
	\begin{equation}
		\label{eq:acc}
		l_m = \int_0^T \frac{1}{2} (\bm{\mu}_m^{\prime T} \bm{P}_{x,m} \bm{\mu}_m^{\prime} - \bm{f}_m(\bm{x})^T \bm{\Pi}_{x,m} \bm{f}_m(\bm{x}) - \bm{\mu}^{\prime T} \bm{P}_x \bm{\mu}^{\prime} + \bm{\eta}_{x}^{\prime T} \bm{\Pi}_x \bm{\eta}_{x}^\prime) dt
	\end{equation}
	where $\bm{\mu}_m$, $\bm{P}_{x,m}$, and $\bm{\Pi}_{x,m}$ are the mean, posterior precision, and prior precision of the $m$th reduced model. In this way, the agent can infer which dynamic hypothesis is most likely to have generated the perceived trajectory -- see Figure \ref{fig:frames_dyn_inf} and \cite{Priorelli2023e} for more details about this approach.
	
	A hybrid unit has useful features deriving from the factorial depths of hidden states and causes. Consider the case where the hidden states encode the agent’s configuration and other environmental objects, while the hidden causes represent the agent’s intentions. A hybrid unit could dynamically assign the causes of its actions at a particular moment: this is critical, e.g., in a pick and place operation, during which an object is first the cause of the hand movements -- resulting in a picking action -- but then it is the consequence of another cause (i.e., a goal position) -- resulting in a placing action. This approach differs from other solutions \cite{Pio-Lopez2016,Friston2010} that directly encode a target location in the hidden causes. Further, embedding environmental entities -- and not just the self -- into the hidden states permits inferring their dynamic trajectories, which is fundamental for interactions, e.g., in catching objects on the fly \cite{Priorelli2023d} or in tracking a hidden target with the eyes \cite{Adams2015b}.
	Considering the example in Figure \ref{fig:frames_dyn_inf}, the hidden states  $\bm{x}$ and $\bm{x}^\prime$ may encode the angle and angular velocity of an agent's arm, and two hidden causes $v_{circle}$ and $v_{square}$ may be associated with dynamics functions $\bm{f}_{circle}$ and $\bm{f}_{square}$ encoding (potential) reaching movements toward the two objects. The environment may be inferred by two likelihood functions, one -- $\bm{g}_p$ -- predicting proprioceptive observations (i.e., arm and angular velocity), and another one -- $\bm{g}_v$ -- predicting visual observations (i.e., the hand position and velocity). Since we are interested in inferring which object is being followed by the agent, we set a uniform discrete prior $\bm{H}_v$. Then, Equation \ref{eq:acc} compares the actual agent's trajectory to the two reaching movements toward the two objects, and assigns a higher probability to the one better resembling the real trajectory.
	
	
	\subsubsection{Hierarchical depth and iterative transformations}
	
\begin{figure}
		\centering
		\begin{minipage}[b]{.48\textwidth}
			\centering
			{\label{fig:deep_kin_graph}\includegraphics[width=\textwidth]{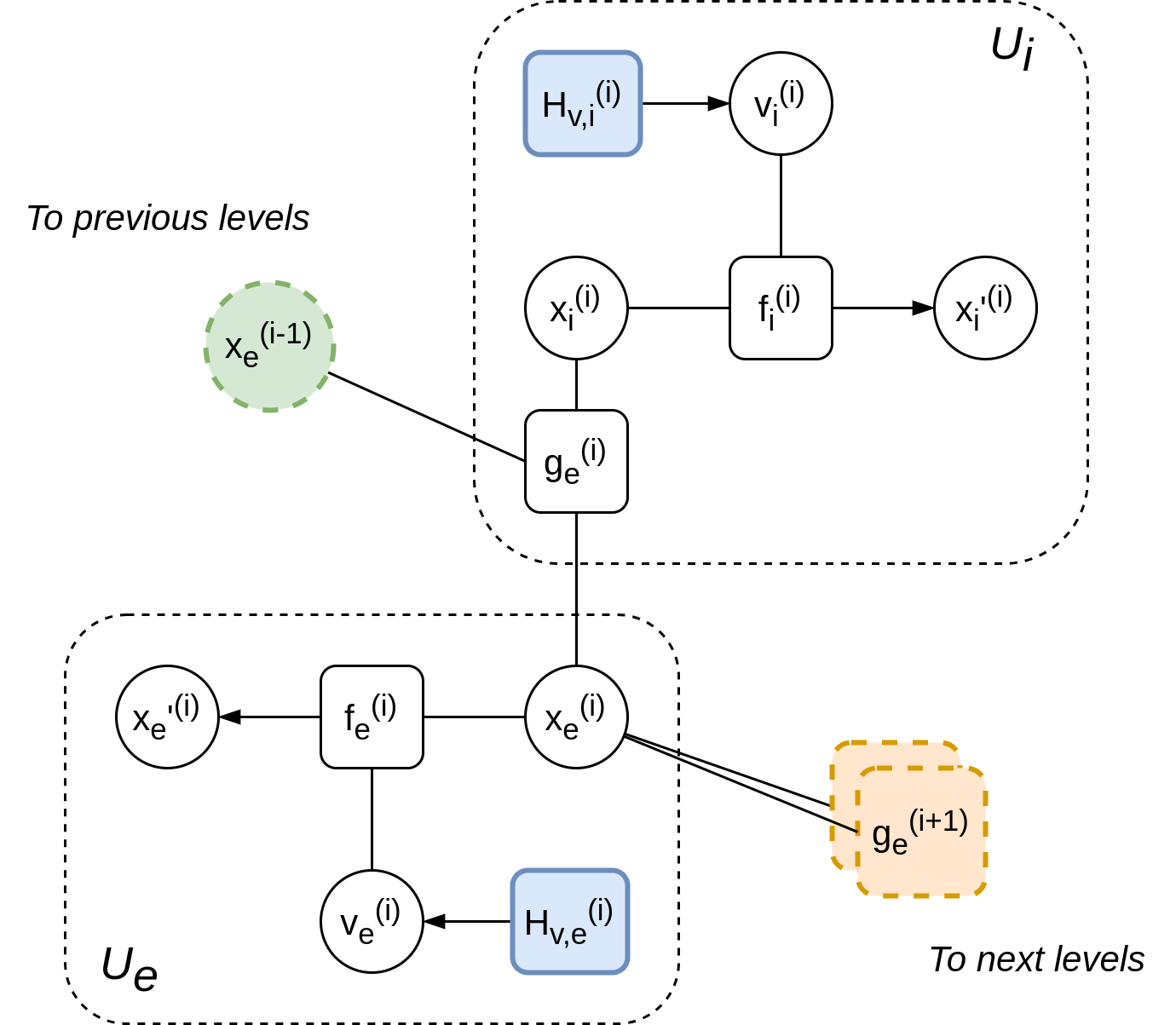}}
		\end{minipage}
		\hfill
		\begin{minipage}[b]{.5\textwidth}
			{\label{fig:example_hier}\includegraphics[width=\textwidth]{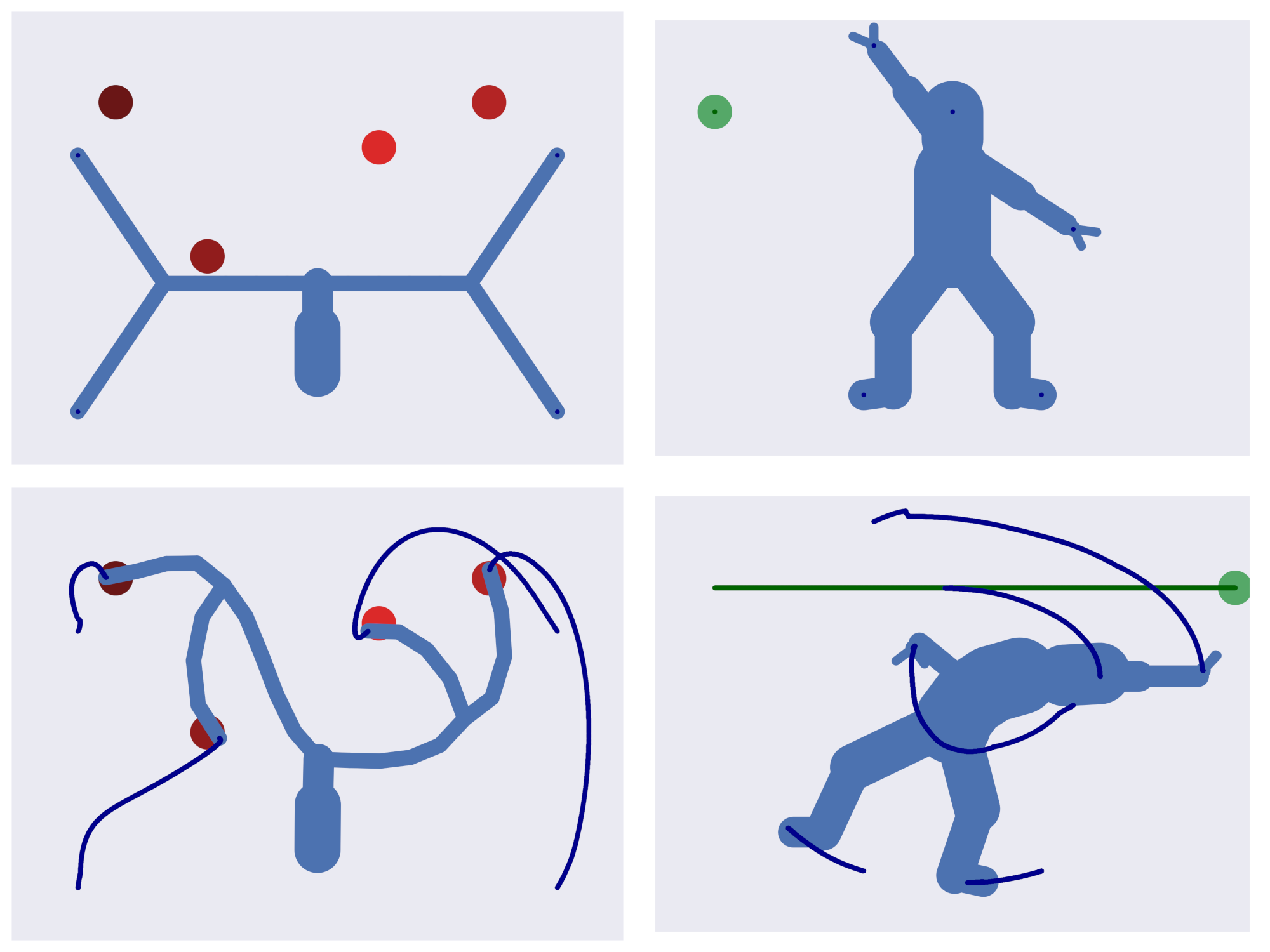}}
		\end{minipage}
		\caption{\textbf{(a)} An IE module is composed of two units $\mathcal{U}_i$ and $\mathcal{U}_e$, which represent a signal in intrinsic and extrinsic reference frames, respectively. Different IE modules can be combined in a hierarchical fashion: the extrinsic signal $\bm{x}_e^{(i)}$ is iteratively transformed through linear transformation matrices encoded in the extrinsic likelihood function $\bm{g}_e^{(i)}$. Hierarchical levels communicate via the 0th-order hidden states. \textbf{(b)} Illustrative examples of a hierarchical model with IE modules. In the first task, the agent (a 23-DoF human body) has to avoid a moving obstacle; in the second task, the agent (a 28-DoF kinematic tree) has to reach four target locations with the extremities of its branches. In both cases, the module in (a) is repeated for every DoF of the agents, matching their kinematic structures. Proprioceptive and exteroceptive (e.g., visual) for each DoF are respectively generated by the intrinsic and extrinsic units via appropriate likelihood functions. See \cite{Priorelli2023b} for more details.}
		\label{fig:deep_kin}
	\end{figure}

	Hierarchical depth is critical in many tasks that require learning of modular and flexible functions. Considering motor control, forward kinematics is repeated throughout every element of the kinematic chain, computing the end effector position from the body-centered reference frame. Iterative transformations are also fundamental in computer vision, where camera models perform roto-translations and perspective projections in sequence. How can we express such hierarchical computations in terms of inference? We design a structure called \textit{Intrinsic-Extrinsic} (or IE) module, performing iterative transformations between reference frames \cite{Priorelli2023b, Priorelli2023c}. A unit $\mathcal{U}_e^{(i)}$ -- where the superscript indicates the $i$th level -- encodes a signal $\bm{x}_e^{(i)}$ in an extrinsic reference frame (e.g., Cartesian coordinates), while another unit $\mathcal{U}_i^{(i)}$ represents an intrinsic signal $\bm{x}_i^{(i)}$ (e.g., polar coordinates). At each level $i$, a likelihood function $\bm{g}_e^{(i)}$ applies a transformation to the extrinsic signal provided by the higher level based on the intrinsic information, and returns a new extrinsic state:
	\begin{equation}
		\label{eq:kin_lkh}
		\bm{x}_e^{(i)} = \bm{g}_e^{(i)}(\bm{x}_i^{(i)}, \bm{x}_e^{(i-1)}) + \bm{w}_e = \bm{T}^{(i)}(\bm{x}_i^{(i)}) \cdot \bm{x}_e^{(i-1)} + \bm{w}_e
	\end{equation}
	where $\bm{w}_e$ is a noise term and $\bm{T}^{(i)}$ is a linear transformation matrix. This new state acts as a prior for the subordinate levels in a multiple-output system; indicating with the superscript $(i,j)$ the $i$th hierarchical level and the $j$th unit within the same level, we link the IE modules in the following way:
	\begin{align}
		\bm{y}^{(i,j)}_e &\equiv \bm{x}^{(i+1,j)}_e &
		\bm{x}^{(i-1,j)}_e &\equiv \bm{\eta}^{(i,j)}_{e}
	\end{align}
	as displayed in Figure \ref{fig:deep_kin};  hence, the observation of level $i$ becomes the prior over the hidden states of level $i+1$. Ill-posed problems that generally have multiple solutions -- such as inverse kinematics or depth estimation -- can be solved by inverting the agent's generative model and backpropagating the sensory prediction errors, with two additional features compared to traditional methods: (i) the possibility of steering the optimization by imposing appropriate priors, e.g., for avoiding singularities during inverse kinematics; (ii) the possibility of acting over the environment to minimize uncertainty, e.g., with motion parallax during depth estimation.
	
	Encoding signals in intrinsic and extrinsic reference frames also induces a decomposition over proprioceptive and exteroceptive predictions, as well as intrinsic and extrinsic dynamics functions, leading to simpler (and yet richer) attractor states.
	Figure \ref{fig:example_hier} shows two examples of goal-directed behavior with complex kinematic structures -- a 28-DoF kinematic tree and a 23-DoF human body. In these cases, an IE module of Figure \ref{fig:deep_kin_graph} is employed for each DoF of the agent. Each IE module encodes the position and velocity of a specific limb, both in an extrinsic (e.g., Cartesian) -- $\bm{x}_e^{(i)}$ and $\bm{x}_e^{(i)\prime}$ -- and intrinsic (e.g., polar) -- $\bm{x}_i^{(i)}$ and $\bm{x}_i^{(i)\prime}$ -- reference frames. These modules are connected hierarchically, i.e., the trunk position generates predictions for the positions of both arms and legs through the likelihood function $\bm{g}_e^{(i)}$ computing forward kinematics. The goal-directed behavior of the kinematic tree is realized by defining four reaching dynamics functions (toward the red objects) at the last levels of the hierarchy representing the end effectors. Instead, the behavior of the human body is achieved by defining repulsive dynamics functions for the extrinsic reference frames of each IE module.
	The decomposition of independent dynamics is also useful for tasks requiring multiple constraints in both domains, e.g., when walking with a glass in hand \cite{Priorelli2023b}, and it has also been applied to controlling robots in 3D environments \cite{Pezzato2024neurips}; or for estimating the depth of an object by moving the eyes \cite{Priorelli2023c}. Further, the factorial depth previously described permits representing hierarchically not only the self, but also the objects in relation to the self, along with the kinematic chains of other agents \cite{Priorelli2025}. In this way, an agent could maintain a \textit{potential body configuration} whenever it observes a relevant entity. This representation also accounts for the affordances of objects to be manipulated, and can be realized efficiently as soon as necessary.
	
	
	\subsubsection{Temporal depth and dynamic planning}
	
	Consider the following discrete generative model:
	\begin{equation}
		p(\bm{s}_{1:\tau},\bm{o}_{1:\tau},\bm{\pi}) = p(\bm{s}_1) p(\bm{\pi}) \prod_{\tau} p(\bm{o}_\tau|\bm{s}_\tau) p(\bm{s}_\tau|\bm{s}_{\tau-1},\bm{\pi})
	\end{equation}
	where:
	\begin{align}
		\begin{aligned}
			p(\bm{s}_1) &= Cat(\bm{D}) \\
			p(\bm{\pi}) &= \sigma(-\bm{\mathcal{G}})
		\end{aligned}
		&&
		\begin{aligned}
			p(\bm{o}_\tau|\bm{s}_\tau) &= Cat(\bm{A}) \\
			p(\bm{s}_{\tau+1} | \bm{s}_\tau, \bm{\pi}) &= Cat(\bm{B}_{\pi,\tau})
		\end{aligned}
	\end{align}
	Here, $\bm{A}$, $\bm{B}$, $\bm{D}$ are the likelihood matrix, transition matrix, and prior, $\bm{\pi}$ is the policy, $\bm{s}_\tau$ are the discrete hidden states at time $\tau$, $\bm{o}_\tau$ are discrete observations, and $\bm{\mathcal{G}}$ is the \textit{expected free energy} (see Section \ref{appendix:hai} for more details).
	
	\begin{figure}
		\centering
		\begin{minipage}[b]{.55\textwidth}
			\centering
			{\label{fig:interface_graph}\includegraphics[width=\textwidth]{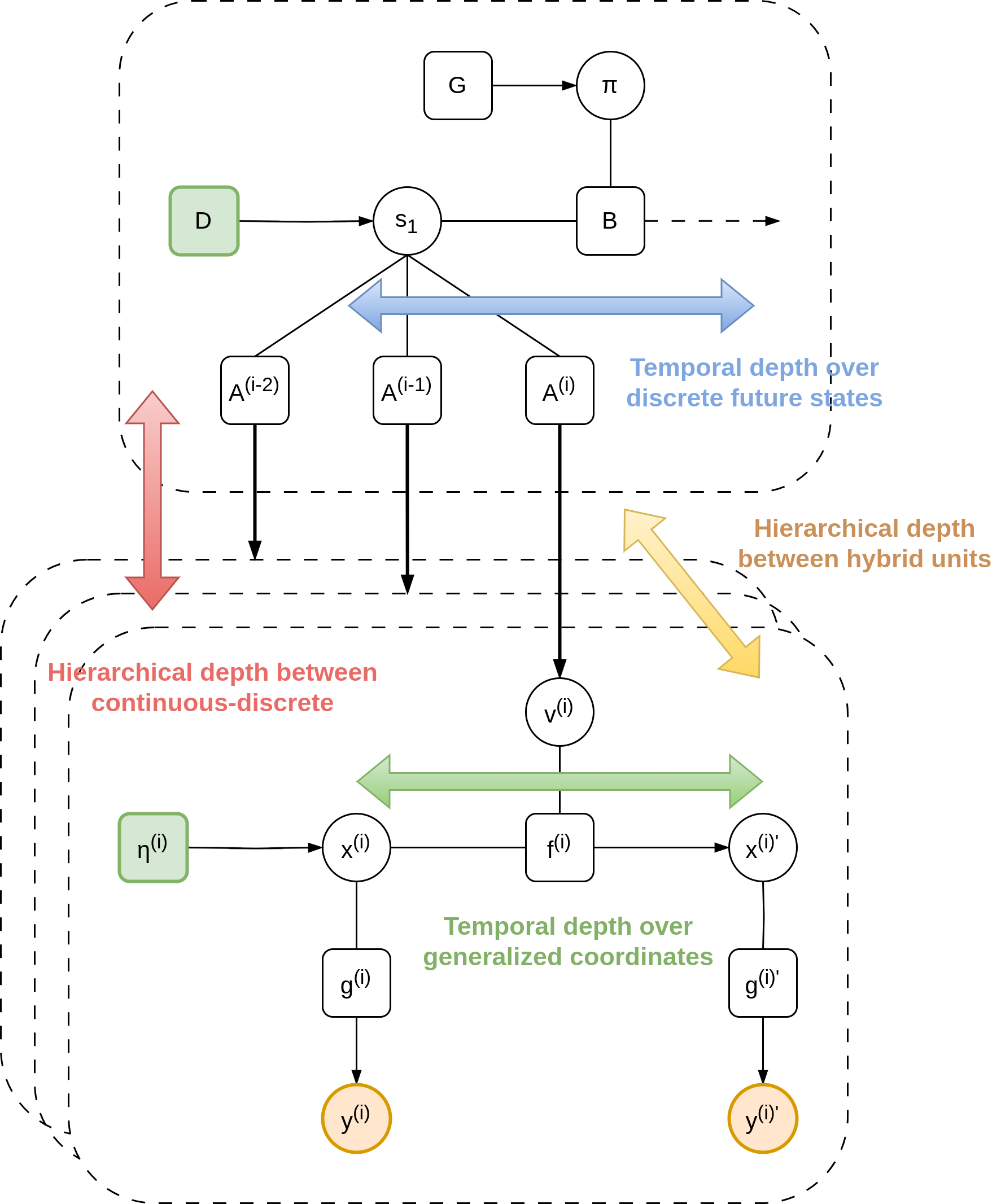}}
		\end{minipage}
		\hfill
		\begin{minipage}[b]{.43\textwidth}
			{\label{fig:example_grasping}\includegraphics[width=\textwidth]{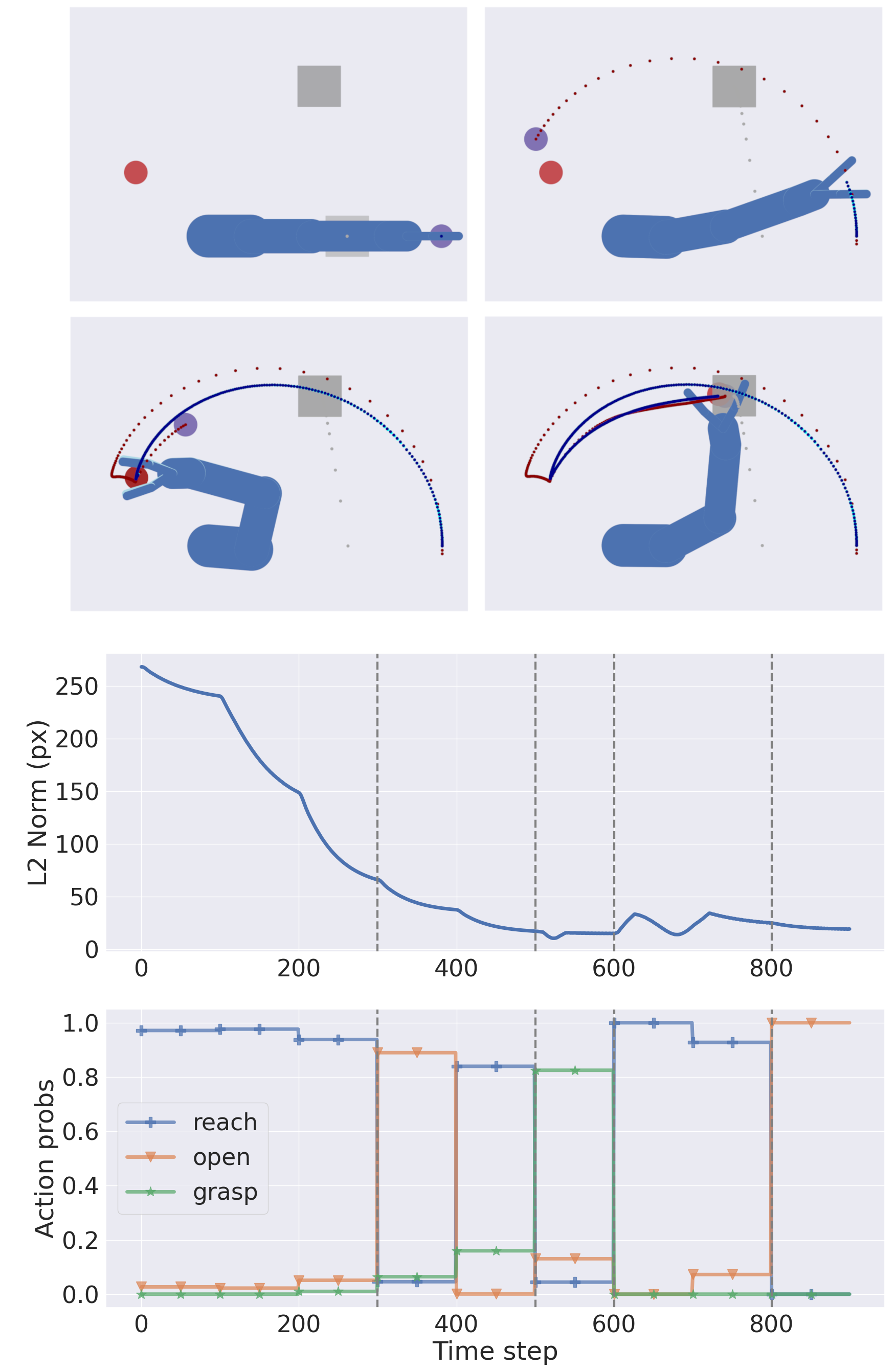}}
		\end{minipage}
		\caption{\textbf{(a)} Interface between a discrete model and several hybrid units. The hidden causes $\bm{v}^{(i)}$ are directly generated, in parallel pathways, from discrete hidden states $\bm{s}_{\tau}$ via likelihood matrices $\bm{A}^{(i)}$. \textbf{(b)} Illustrative example with the hybrid units combined with a discrete model. In this task, the agent (a 4-DoF arm with an additional 4-DoF hand composed of two fingers) has to pick a moving ball (the red circle) and place it at a goal position (the grey square). The discrete hidden states $\bm{s}_{\tau}$ encode the agent position (start position, at the ball, or at the goal) and the status of the hand (open or closed). These are informed by two continuous models encoding intrinsic (joint angles) and extrinsic (hand and objects positions) information, respectively. The hidden causes $\bm{v}$ of the intrinsic model are related to hand opening and closing actions, while the hidden causes of the extrinsic model relate to two reaching movements, as in the previous case. Note that the object belief (purple circle) is rapidly inferred, and as soon as the picking action is complete, the belief is gradually pulled toward the goal position, resulting in a second reaching movement. The top right panel shows the hand-object distance over time, while the bottom right panel displays the dynamics of the discrete action probabilities used to infer the next discrete state. The vertical dashed lines distinguish five different phases: a pure reaching movement, an intermediate phase when the agent prepares the grasping action, a grasping phase, a second reaching movement and, finally, the ball release. The stepped behavior of the action probabilities is due to the replanning made by the discrete model every $10$ continuous time steps. See \cite{Priorelli2023d} for more details.}
		\label{fig:interface}
	\end{figure}
	
	We can let the likelihood $p(\bm{o}_\tau | \bm{s}_\tau)$ directly bias the (discrete) hidden causes of a hybrid unit:
	\begin{align}
		\bm{H}_\tau &\equiv \bm{A} \bm{s}_{\tau} &
		\bm{o}_\tau &\equiv \bm{v}_\tau
	\end{align}
	Hence, the prior $\bm{H}_\tau$ over the discrete hidden causes becomes the prediction by the discrete model, while the discrete observation becomes the discrete hidden causes of the hybrid unit. This is an alternative method to the state-of-the-art, which considers an additional level between discrete observations and static priors over continuous hidden causes \cite{Friston2017}. Here, the discrete model can impose priors over trajectories even in the same period $\tau$, thus affording dynamic planning \cite{Priorelli2023e,Priorelli2025}. If a discrete model is linked to different hybrid units in parallel -- as shown in Figure \ref{fig:interface} -- the discrete hidden states are inferred by combining multiple evidences:
	\begin{align}
		\begin{split}
			\label{eq:dyn_plan}
			\bm{s}_{\pi,\tau} &= \sigma( \ln \bm{B}_{\pi,\tau-1} \bm{s}_{\pi,\tau-1} + \bm{B}_{\pi,\tau+1}^T \bm{s}_{\pi,\tau+1} + \sum_n \ln \bm{A}^{(i)^T} \bm{v}^{(i)}_{\tau} )
		\end{split}
	\end{align}
	where $\bm{s}_{\pi,\tau}$ are the discrete hidden states conditioned over policy $\pi$ at time $\tau$, while the superscript $i$ indicates the $i$th hybrid unit. These parallel pathways synchronize the behavior of all low-level units based on the same high-level plan, allowing, e.g., simultaneous coordination of every limb of the human body. 
	
	In Figure \ref{fig:interface_graph}, we notice the two kinds of \textit{temporal} depths, peculiar to hybrid active inference. The first one comes from the discrete component and unfolds over future states $(\bm{s}_1, \bm{s}_2, \bm{s}_3, \dots)$ over a time horizon defined by the policy length: it allows the agent to make plans by computing the expected free energy of those states \cite{Friston2021a}. The second temporal depth derives from the continuous level and unfolds over the temporal derivatives of the hidden states, i.e., $(\bm{x}, \bm{x}^\prime, \bm{x}^{\prime\prime}, \dots)$: this refines the estimated trajectories with an increasing sampling rate.
	
	In addition to this, the overall deep hybrid model presents two \textit{hierarchical} depths with different roles. First, a hybrid scale that separates the slow-varying representation of the discretized task with the fast update of continuous signals. Here, the temporal predictions from the continuous dynamics are used to infer accurately the discrete variables, so that the agent can make complex high-level plans even when the surrounding environment is changing frequently, and is able to revise those plans when new evidence has been accumulated. Second, a continuous scale linking the hybrid units and inherent to the hierarchical representation of the agent's kinematic structure, which can be appreciated from Figure \ref{fig:deep_kin}. This induces a separation of temporal scales between high and low levels of the hierarchy (e.g., the trunk vs the hands), as the predictions errors generated from the dynamics of the hand have a less and less impact as they flow back to the shoulder and trunk dynamics.
	
	Considering the pick-and-place operation shown in Figure \ref{fig:example_grasping}, we can encode the three key moments of the task (start position, ball picked, and ball placed) in terms of discrete hidden states $\bm{s}$. At each discrete step $\tau$, these states make (intrinsic and extrinsic) predictions for the hybrid units representing the agent's kinematic chain (as in Figure \ref{fig:deep_kin}). For instance, the second discrete hidden state generates a potential body configuration with the hand closed and at the ball position. We can use an identity mapping for the likelihood matrices, so that the intrinsic and extrinsic hidden causes of the hybrid units all have the same decomposition into three steps -- i.e., $\bm{v}^{(i)}_i = [v^{(i)}_{i, start}, v^{(i)}_{i, picked}, v^{(i)}_{i, placed}]$ and $\bm{v}^{(i)}_e = [v^{(i)}_{e, start}, v^{(i)}_{e, picked}, v^{(i)}_{e, placed}]$. Notably, since these hidden causes are related to potential trajectories, the agent can pick and place the ball even in dynamic contexts, e.g., if the ball is moving.
	
	
	\subsection{A deep hybrid model for tool use}
	
	In this section, we show how a deep hybrid model can be used efficiently in a task that requires planning in a dynamic environment and coordination of all elements of the agent's body. The implementation details are found hereafter in Section \ref{appendix:impl}, while Appendix A illustrates the algorithms for the inference of the discrete model and hybrid units. Then, in Section \ref{analysis} we analyze model performance and describe the effects of dynamic planning in terms of accumulated sensory evidence and transitions over discrete hidden states.
	
	\begin{figure}
		\centering
		\begin{minipage}[b]{.2\textwidth}
			\centering
			{\label{fig:env}\includegraphics[width=\textwidth]{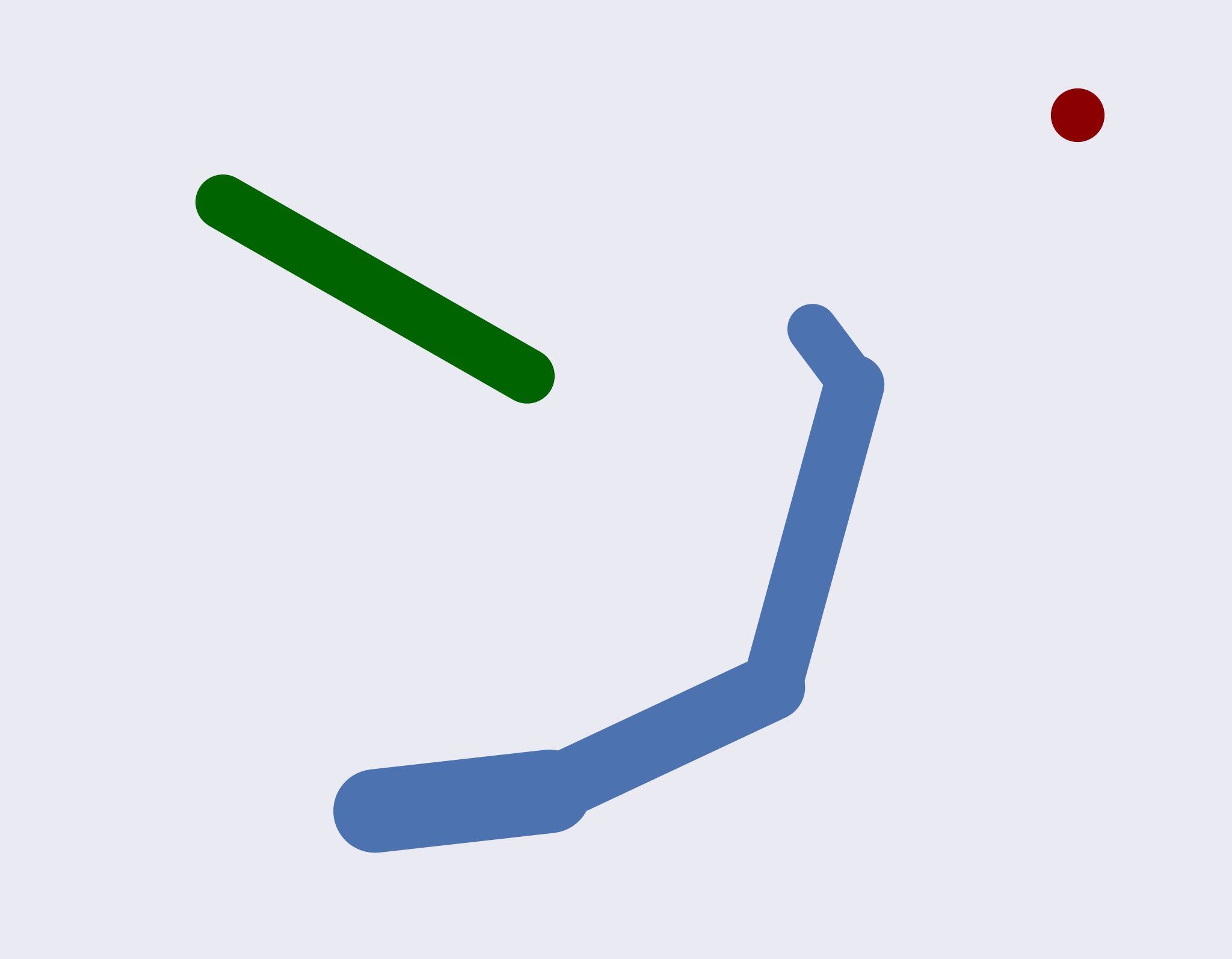}}
			\vspace{2em}
			{\label{fig:env_kin}\includegraphics[width=\textwidth]{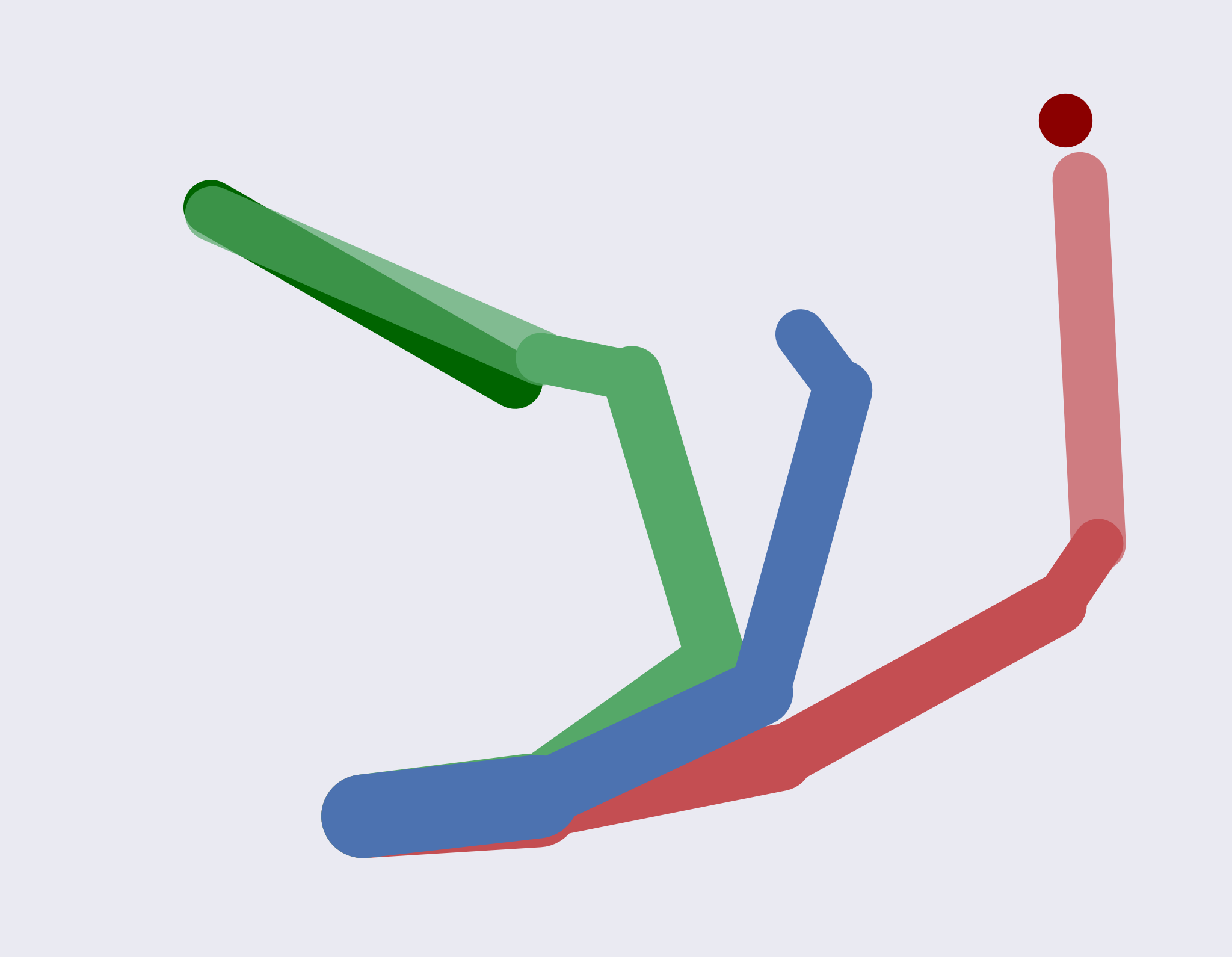}}
		\end{minipage}
		\hfill
		\begin{minipage}[b]{.77\textwidth}
			{\label{fig:paths}\includegraphics[width=\textwidth]{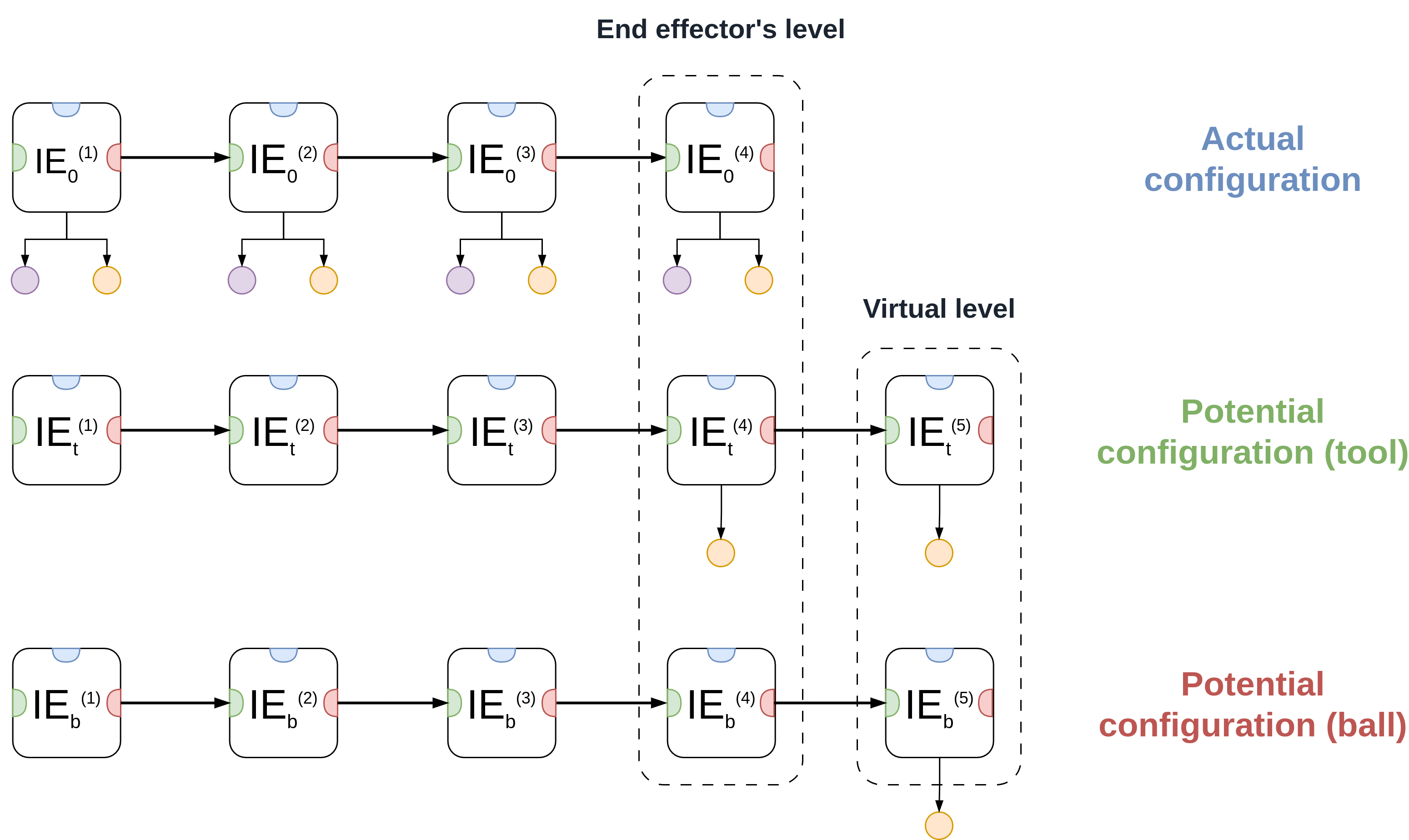}}
		\end{minipage}
		\caption{\textbf{(a)} Virtual environment of the tool use task. An agent controlling a 4-DoF arm has to grasp a moving tool (in green) and reach a moving ball (in red) with the tool's extremity. \textbf{(b)} Agent's beliefs over the continuous hidden states of the arm (blue), tool (light green), and ball (light red). The real positions of the tool and ball are represented in dark green and dark red, respectively. The virtual level is plotted with more transparent colors. \textbf{(c)} Graphical representation of the agent's continuous generative model. Every environmental entity is encoded hierarchically by considering the whole arm's kinematic structure. For clarity, the three pathways are displayed separately, while lateral connections and the high-level discrete model are not shown. The end effector's level encodes intrinsic and extrinsic information about the end effector, regarding the three configurations (the actual end effector position, the belief over the end effector at the tool's origin, or at an appropriate position to reach the ball with the tool's extremity). Instead, the virtual level is not present in the actual configuration, since the tool is not part of the agent's kinematic chain and it is only used in the generative model for goal-directed behavior -- as if it were a new joint. This level encodes intrinsic and extrinsic information about the tool, regarding the two potential configurations (the belief over the tool's extremity at the actual tool's extremity, and at the actual ball position.) Small purple and yellow circles represent proprioceptive and exteroceptive observations, respectively.}
	\end{figure}
	
	Reaching an object with a tool is a complex task that requires all the features delineated in the previous section. First, the task has to be decomposed into subgoals -- reaching the tool and reaching the object -- which requires high-level discrete planning. Second, the agent has to maintain distinct beliefs about its arm, the tool, and the object, all of which must be inferred from sensory observations if their locations are unknown or constantly changing. Third, if the tool has to be grasped at the origin while the object has to be reached with the tool's extremity, the agent's generative model should encode a hierarchical representation of the self and every entity, and specify goals (in the form of attractors) at different levels of the hierarchy.
	
	As shown in the graphical representation of the virtual environment of Figure \ref{fig:env}, the agent controls an arm of 4 DoF. The agent receives proprioceptive information about its joint angles, and exteroceptive (e.g., visual) observations encoding the positions of its limbs, the tool, and the ball. For simplicity, we assume that the tool sticks to the agent's end effector as soon as it is touched. The generative model provides an effective decomposition into three parallel pathways, displayed in Figure \ref{fig:paths}: one maintaining an estimate of the agent's \textit{actual} configuration (indicated by the subscript 0), and two others representing \textit{potential} configurations \textit{in relation to} the tool and the ball (respectively indicated by the subscripts t and b). In other words, the objects of interest are not just encoded by their qualities or location, but already define a body configuration appropriate to achieve a specific interaction. In our case, the potential configuration related to the tool represents not only the estimated tool's location, but also the estimated end effector's location needed to reach the tool. In addition, each body configuration is composed of as many IE modules as the agent's DoF, following the hierarchical relationships of the forward kinematics and allowing to express both joint angles and limb positions.
	Message passing of \textit{extrinsic prediction errors}, i.e., differences between the estimated limb positions and their predictions given by the estimated joint angles, allows to infer the whole body configuration (either actual or potential) via exteroceptive observations. In addition, every component of a level exchanges lateral messages with the other components, in the form of \textit{dynamics prediction errors}. These errors are caused by the potential dynamics functions described in Section \ref{section:flexible}, which define the interactions between entities for goal-directed behavior.
	
	\begin{figure}
		\centering
		\includegraphics[width=\textwidth]{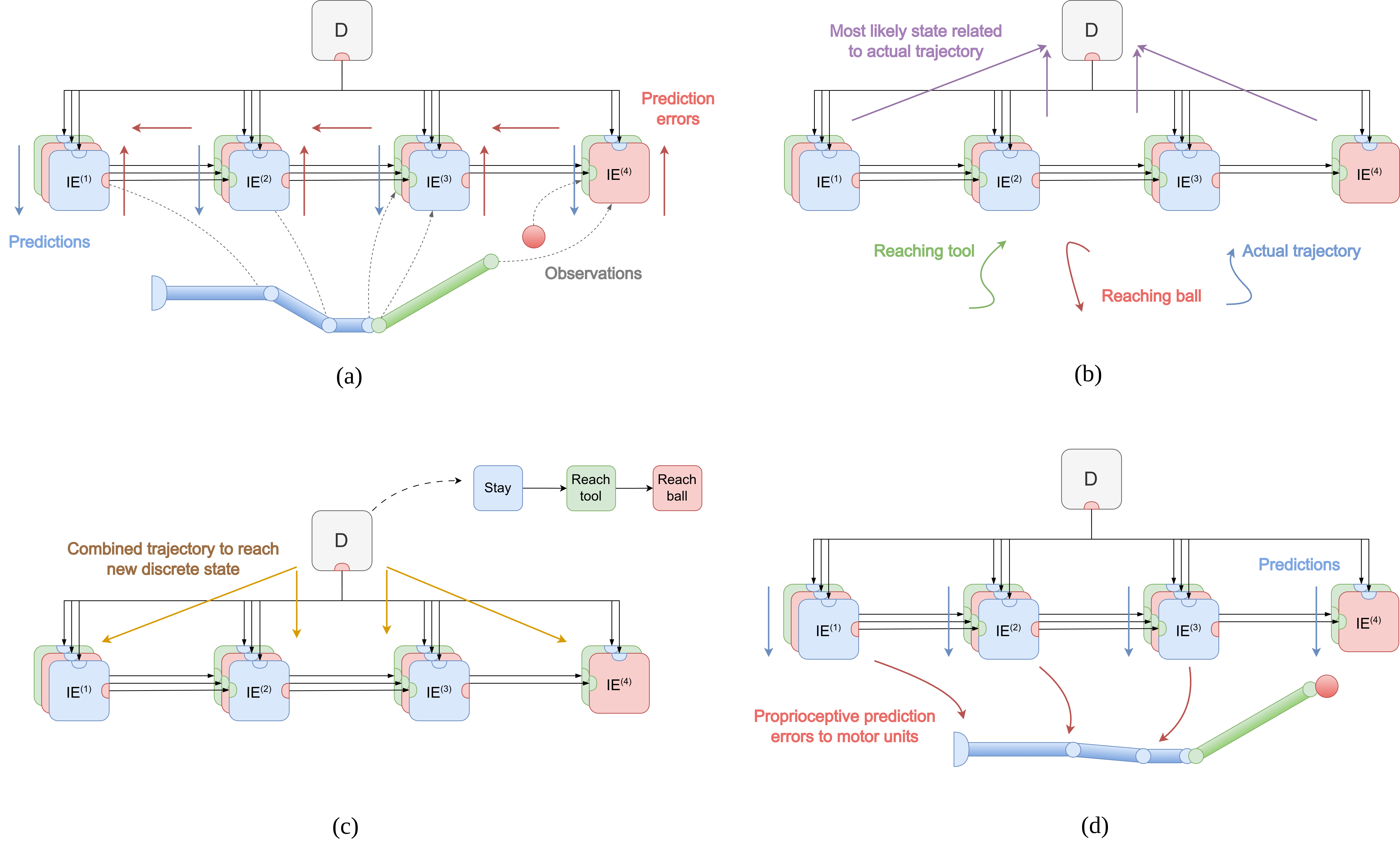}
		\caption{Graphical representation of a deep hybrid model for tool use, composed of a discrete model at the top and several IE modules. Every module is factorized into three elements, related to the observations of the agent's arm (in blue), a tool (in green), and a ball (in red). Note that the last (virtual) level only considers the tool's extremity and the ball. The computation of the action for a single time step is divided into four main processes. (a) \textbf{Perception}. Proprioceptive and visual observations $\bm{y}_p$ and $\bm{y}_e$ are compared with the agent's predictions. The resulting prediction errors are propagated throughout the hierarchy to infer the actual kinematic configuration, as well as potential configurations related to the objects. (b) \textbf{Dynamic inference}. The bottom-up messages $\bm{l}_e$ from the IE modules inform the discrete model about the most likely state that may have generated the perceived arm trajectory. This is done by comparing the latter with potential trajectories $\bm{f}_m$ related to dynamic hypotheses $\bm{v}_e$ (see Equation \ref{eq:causes}). For instance, if the agent is reaching the tool and the ball is moving away, the bottom-up messages assign a higher probability to the tool-reaching hypothesis and a lower probability to the initial steady state. (c) \textbf{Dynamic planning}. The agent infers the next discrete action to take by minimizing the expected free energy $\mathcal{G}$ (see Equation \ref{eq:exp_fe}). As a result, the agent believes to be at the next discrete state, corresponding to the ball-reaching hypothesis. In turn, this biased state generates a new combined trajectory (through the discrete extrinsic prediction $\bm{A}_e \bm{s}$ in Equation \ref{eq:causes}), acting as a prior for the continuous hidden states of the IE modules. (d) \textbf{Action}. The continuous hidden states generate predictions, which are again compared with the related observations. The proprioceptive prediction errors climb back the hierarchy as before, but they are also suppressed through movement by motor units (see Equation \ref{eq:action}). This second process eventually produces a continuous action that moves the end effector toward the ball.}
		\label{fig:all}
	\end{figure}
	
	Two crucial aspects arise when modeling entities in a deep hierarchical fashion and related to the self. First, different configurations are inferred (hence, different movements) depending on the desired interaction with the object considered -- for example, reaching a ball with either the elbow or the end effector. Second, entities could have their own hierarchical structures: in our application, the tool consists of two Cartesian positions and an orientation, and the agent should somehow represent this additional link. For these reasons, we consider a virtual level for the tool configuration, attached to the last IE module (i.e., end effector), as exemplified in Figure \ref{fig:paths}; the visual observations of the tool are then linked to the last two levels. From these observations, the correct tool angle can be inferred as if it were a new joint angle of the arm. Additionally, since we want the agent to touch the ball with the tool's extremity, we model the third (ball) configuration with a similar structure, in which a visual observation of the ball is attached to the virtual level. The overall architecture can be better understood from Figure \ref{fig:env_kin}, showing the agent's continuous beliefs of all three entities. As soon as the agent perceives the tool, it infers a possible kinematic configuration as if it had visual access only to its last two joints (which are actually the tool's origin and extremity). Likewise, perceiving the ball causes the agent to find an extended kinematic configuration as if the tool were part of the arm.
	
	\begin{figure}
		\centering
		\includegraphics[width=0.95\textwidth]{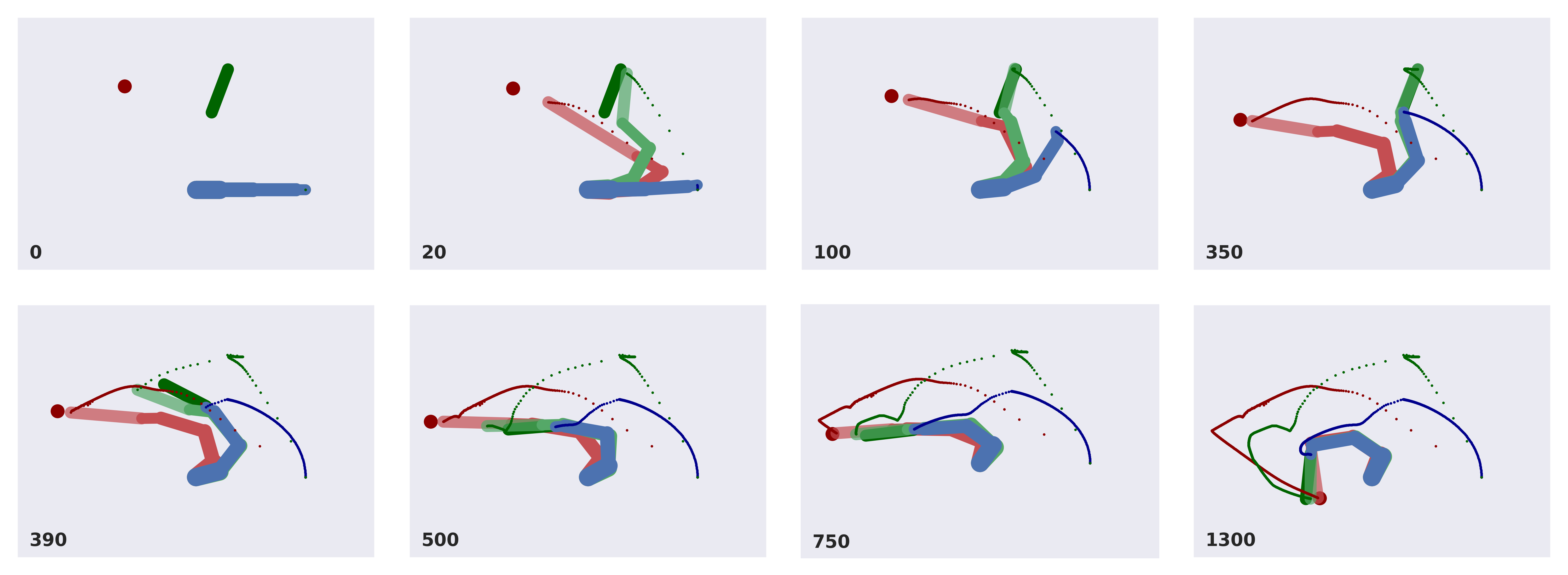}
		\caption{Sequence of time frames of the simulation. Real ball, tool, and arm are displayed in dark red, dark green, and dark blue respectively. Beliefs of tool and ball, in terms of potential kinematic configurations, are shown in light green and light red, respectively. Trajectories of the end effector, tool, and ball are displayed as well. The number of time steps is shown in the lower-left corner of each frame.}
		\label{fig:frames}
	\end{figure}
	
	With this task formalization, specifying the correct dynamics for goal-directed behavior is simple. First, we define two sets of dynamics functions implementing every subgoal, one for reaching the tool and another for reaching the ball with the tool's extremity. As explained in Section \ref{appendix:impl}, the second subgoal requires specifying an attractor at the virtual level, which makes the agent think that the tool's extremity will be pulled toward the ball. The biased state generates an extrinsic prediction error that is backpropagated to the previous level encoding the tool's origin and the end effector. Notably, defining discrete hidden states and discrete hidden causes related to the agent's intentions allows the agent to accumulate evidence over trajectories from different modalities (e.g., intrinsic or extrinsic) and hierarchical locations (e.g., elbow or end effector), ultimately solving the task via inference. The four main processes of the task -- i.e., \textit{perception}, \textit{dynamic inference}, \textit{dynamic planning}, and \textit{action} -- are summarized in Figure \ref{fig:all}.
	
	\subsubsection{\label{appendix:impl}Implementation details}
	
	The agent's sensory modalities are: (i) a proprioceptive observation $\bm{y}_p$ for the arm's joint angles; (ii) a visual observation $\bm{y}_v$ encoding the Cartesian positions of every link of the arm, both extremities of the tool, and the ball; (iii) a discrete tactile observation $\bm{o}_t$ signaling whether or not the target is grasped.
	
	We decompose intrinsic and extrinsic hidden states of every IE module into three components, the first one corresponding to the actual arm configuration, and the other two related to potential configurations for the tool and the ball. Hence:
	\begin{align}
		\begin{split}
			\bm{x}_i^{(i)} &= \begin{bmatrix}
				\bm{x}^{(i)}_{i,0} & \bm{x}^{(i)}_{i,t} & \bm{x}^{(i)}_{i,b}
			\end{bmatrix} \\
			\bm{x}_e^{(i)} &= \begin{bmatrix}
				\bm{x}^{(i)}_{e,0} & \bm{x}^{(i)}_{e,t} & \bm{x}^{(i)}_{e,b}
			\end{bmatrix}
		\end{split}
	\end{align}
	The end effector's level and the virtual level (see Figure \ref{fig:paths}) are indicated with the superscripts $(4)$ and $(5)$, respectively. Regarding the IE module of the virtual level, the intrinsic and extrinsic hidden states only have two components related to the potential states of the tool and the ball, i.e., $\bm{x}_i^{(5)} = [\bm{x}^{(5)}_{i,t}, \bm{x}^{(5)}_{i,b}]$ and $\bm{x}_e^{(5)} = [\bm{x}^{(5)}_{e,t}, \bm{x}^{(5)}_{e,b}]$.
	
	For each entity, the intrinsic hidden states encode pairs of joint angles and limb lengths, e.g., $\bm{x}_{i,0}^{(i)} = [\theta_0^{(i)},l_0^{(i)}]$ while the extrinsic reference frame is expressed in terms of the position of a limb's extremity and its absolute orientation, e.g., $\bm{x}_{e,0}^{(i)} = [p_{0,x}^{(i)},p_{0,y}^{(i)},\phi_0^{(i)}]$. The likelihood function $\bm{g}_e$ of Equation \ref{eq:kin_lkh} computes extrinsic predictions independently for each entity:
	\begin{equation}
		\bm{g}_e(\bm{x}_i^{(i)},\bm{x}_e^{(i-1)}) = \begin{bmatrix} \bm{T}(\bm{x}^{(i)}_{i,0}, \bm{x}^{(i-1)}_{e,0}) & \bm{T}(\bm{x}^{(i)}_{i,t}, \bm{x}^{(i-1)}_{e,t}) & \bm{T}(\bm{x}^{(i)}_{i,b}, \bm{x}^{(i-1)}_{e,b}) \end{bmatrix}
	\end{equation}
	Here, the mapping $\bm{T}(\bm{x}_i, \bm{x}_e)$ reduces to a simple roto-translation:
	\begin{equation}
		\bm{T}(\bm{x}_i, \bm{x}_e) = \begin{bmatrix} p_x + l c_{\theta,\phi} \\ p_y + l s_{\theta,\phi} \\ \phi + \theta \end{bmatrix}
	\end{equation}
	where $\bm{x}_{i} = [\theta,l]$, $\bm{x}_{e} = [p_{x},p_{y},\phi]$, and we used a compact notation to indicate the sine and cosine of the sum of two angles, i.e., $c_{\theta,\phi} = \cos(\theta) \cos(\phi) - \sin(\theta) \sin(\phi)$.
	Each level then computes proprioceptive and visual predictions through likelihood functions $\bm{g}_p$ and $\bm{g}_v$, which in this case are simple mappings that extract the joint angles of the actual arm configuration and the Cartesian positions of the limbs and objects from the intrinsic and extrinsic hidden states, respectively:
	\begin{align}
		\begin{split}
			\bm{g}_p(\bm{x}_i^{(i)}) &= \theta_0^{(i)} \\
			\bm{g}_v(\bm{x}_e^{(i)}) &= \begin{bmatrix}
				p_{0,x}^{(i)} & p_{t,x}^{(i)} & p_{b,x}^{(i)} \\ p_{0,y}^{(i)} & p_{t,y}^{(i)}  & p_{b,y}^{(i)}
			\end{bmatrix}
		\end{split}
	\end{align}
	
	\begin{figure}
		\centering
		\begin{minipage}[b]{.35\textwidth}
			\centering
			{\label{fig:tool_ints_1st}\includegraphics[width=\textwidth]{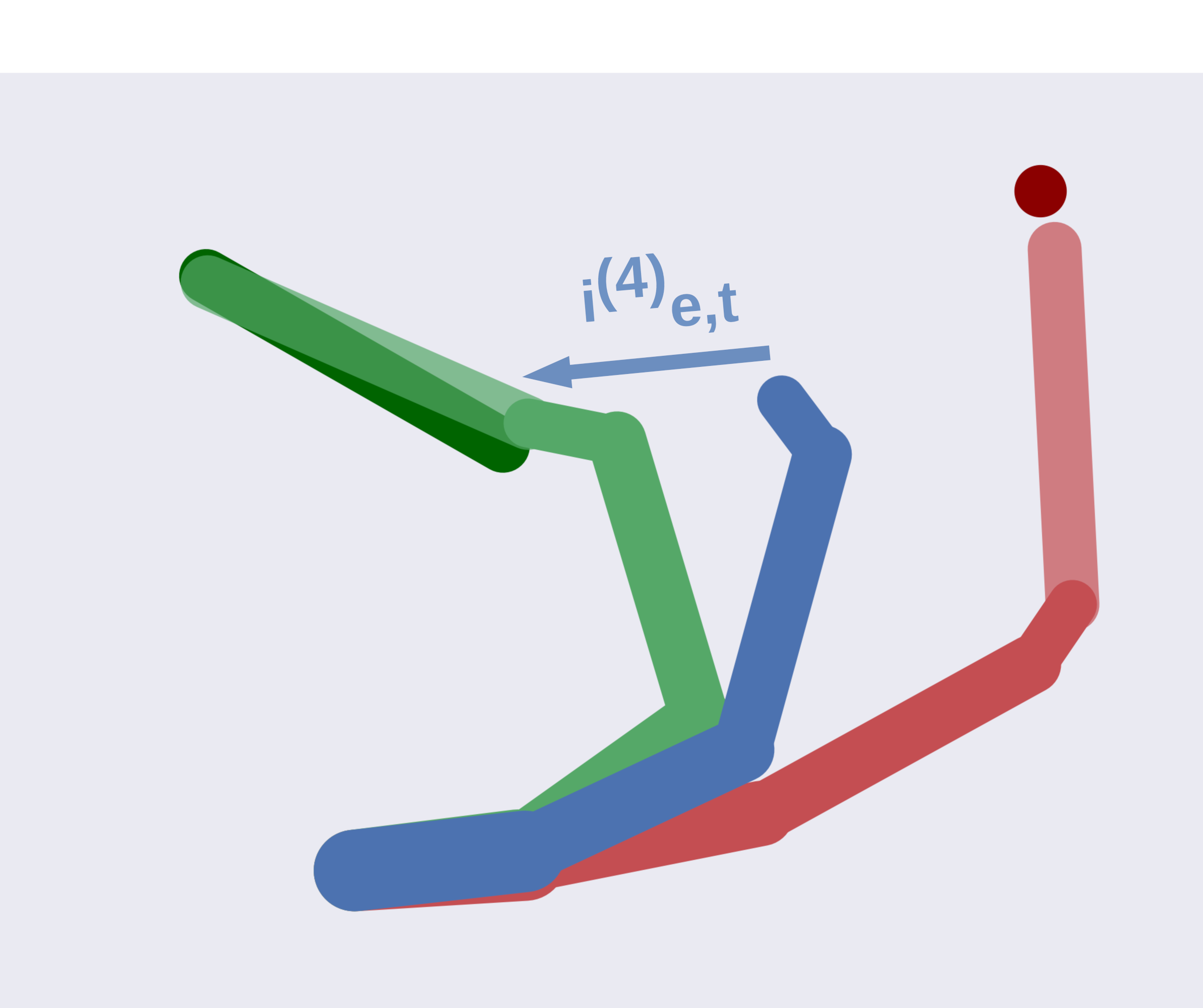}}
		\end{minipage}
		\hfill
		\begin{minipage}[b]{.35\textwidth}
			{\label{fig:tool_ints_2nd}\includegraphics[width=\textwidth]{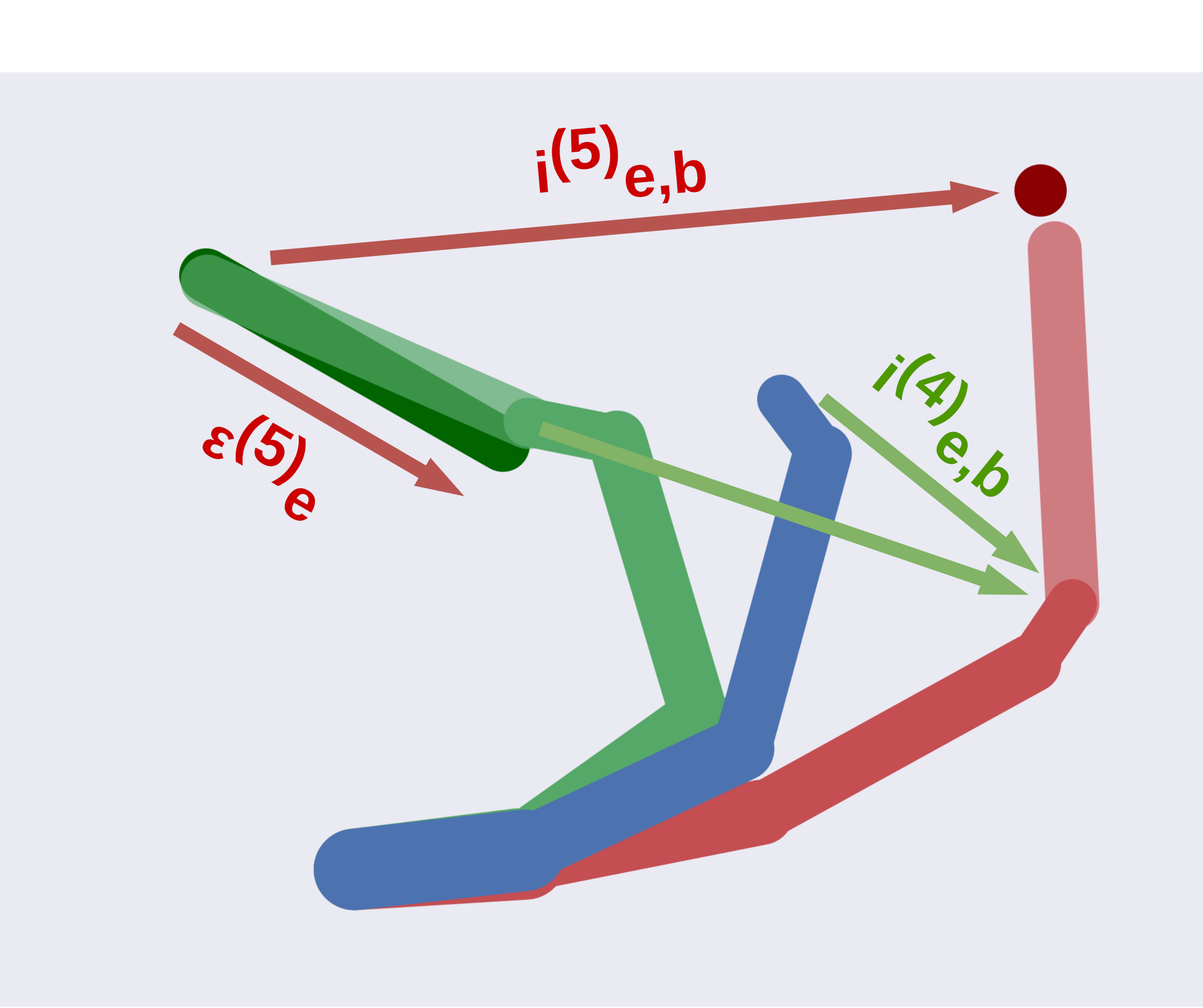}}
		\end{minipage}
		\caption{Representation of the dynamics active during the two steps. \textbf{(a)} The end effector is pulled toward the belief about the tool's origin through $\bm{f}^{(4)}_{e,t}$. \textbf{(b)} The tool's extremity is pulled toward the ball through dynamics $\bm{f}^{(5)}_{e,b}$. This generates an extrinsic prediction error $\bm{\varepsilon}^{(5)}_e$ that steers the previous level of the potential configuration of the tool. Concurrently, a second dynamics $\bm{f}^{(4)}_{e,b}$ also pulls both actual and tool components of the end effector's level toward the potential configuration of the ball.}
		\label{fig:tool_ints}
	\end{figure}
	
	Reaching the tool's origin with the end effector is achieved by a function (related to an agent's intention) that sets the first component of the corresponding extrinsic hidden states equal to the second one:
	\begin{equation}
		\bm{i}^{(4)}_{e,t}(\bm{x}^{(4)}_e) = \begin{bmatrix} \bm{x}^{(4)}_{e,t} & \bm{x}^{(4)}_{e,t} & \bm{x}^{(4)}_{e,b} \end{bmatrix}
	\end{equation}
	Then, we define a \textit{potential dynamics function} by subtracting the current hidden states from this intentional state:
	\begin{align}
		\begin{split}
			\bm{f}_{e,t}^{(4)}(\bm{x}^{(4)}_e) &= \bm{i}^{(4)}_{e,t}(\bm{x}^{(4)}_e) - \bm{x}^{(4)}_e = \begin{bmatrix} \bm{x}^{(4)}_{e,t} - \bm{x}^{(4)}_{e,0} & \bm{0} & \bm{0} \end{bmatrix}
		\end{split}
	\end{align}
	Note the decomposition into separate attractors. A non-zero velocity for the first component expresses the agent's desire to move the end effector, while a zero velocity for the other two components means that the agent does not intend to manipulate the objects during the first step of the task. Since a potential kinematic configuration for the tool is already at the agent's disposal, in order to speed up the movement similar functions can be defined at every hierarchical level, both in intrinsic and extrinsic reference frames. The second step of the task involves reaching the ball with the tool's extremity. Hence, we define two intentional states for the end effector's and virtual levels, setting every component equal to the (potential) component related to the ball:
	\begin{align}
		\begin{split}
			\bm{i}^{(4)}_{e,b}(\bm{x}^{(4)}_e) &= \begin{bmatrix} \bm{x}^{(4)}_{e,b} & \bm{x}^{(4)}_{e,b} & \bm{x}^{(4)}_{e,b} \end{bmatrix} \\
			\bm{i}^{(5)}_{e,b}(\bm{x}^{(5)}_e) &= \begin{bmatrix} \bm{x}^{(5)}_{e,b} & \bm{x}^{(5)}_{e,b} \end{bmatrix}
		\end{split}
	\end{align}
	The attractors encoded in the second set of potential dynamics functions express the agent's desire to modify the tool's location:
	\begin{align}
		\begin{split}
			\bm{f}_{e,b}^{(4)}(\bm{x}^{(4)}_e) &= \bm{i}^{(4)}_{e,b}(\bm{x}^{(4)}_e) - \bm{x}^{(4)}_e = \begin{bmatrix} \bm{x}^{(4)}_{e,b} - \bm{x}^{(4)}_{e,0} & \bm{x}^{(4)}_{e,b} - \bm{x}^{(4)}_{e,t} & \bm{0} \end{bmatrix} \\
			\bm{f}_{e,b}^{(5)}(\bm{x}^{(5)}_e) &= \bm{i}^{(5)}_{e,b}(\bm{x}^{(5)}_e) - \bm{x}^{(5)}_e = \begin{bmatrix} \bm{x}^{(5)}_{e,b} - \bm{x}^{(5)}_{e,t} & \bm{0} \end{bmatrix}
		\end{split}
	\end{align}
	Maintaining this set of dynamics eventually drives the hand into a suitable position that makes the tool's extremity touch the ball. A representation of the relations between such dynamics is displayed in Figure \ref{fig:tool_ints}.
	
	Now, we define the hidden causes of the last two levels (for simplicity, we only describe the extrinsic hidden states):
	\begin{align}
		\begin{split}
			\bm{v}_e^{(4)} &= \begin{bmatrix} v_s^{(4)} & v_t^{(4)} & v_b^{(4)} \end{bmatrix} \\
			\bm{v}_e^{(5)} &= \begin{bmatrix} v_s^{(5)} & v_b^{(5)} \end{bmatrix}
		\end{split}
	\end{align}
	where the subscripts $s$, $t$, and $b$ indicate the agent's intentions to maintain the current state of the world ("\textit{stay}"), reach the tool, and reach the ball, respectively. The first hidden cause, related to the following intentional states:
	\begin{align}
		\bm{i}_{e,s}^{(4)}(\bm{x}^{(4)}_e) &= \bm{x}^{(4)}_e &
		\bm{i}_{e,s}^{(5)}(\bm{x}^{(5)}_e) &= \bm{x}^{(5)}_e
	\end{align}
	are needed to ensure that $\bm{v}_e^{(4)}$ and $\bm{v}_e^{(5)}$ encode proper probabilities when the discrete model is in the initial state. The average trajectory is then found by weighting the potential dynamics functions with the corresponding hidden causes. Hence, having indicated with $\bm{\mu}_e^{(4)}$ the \textit{belief} of the extrinsic hidden states of the end effector, the generated trajectory is:
	\begin{align}
		\begin{split}
			\bm{\eta}_{x,e}^{\prime(4)} &= v_{s}^{(4)} \bm{f}_{e,s}^{(4)}(\bm{\mu}_e^{(4)}) + v_{t}^{(4)} \bm{f}_{e,t}^{(4)}(\bm{\mu}_e^{(4)}) + v_{b}^{(4)} \bm{f}_{e,b}^{(4)}(\bm{\mu}_e^{(4)})
		\end{split}
	\end{align}
	The belief is then updated according to the following update rules:
	\begin{align}
		\begin{split}
			\label{eq:update}
			\dot{\bm{\mu}}_e^{(4)} &= \bm{\mu}^{\prime(4)}_e - \bm{\pi}_e^{(4)} \bm{\varepsilon}_e^{(4)} + \partial \bm{g}_e^T \bm{\pi}_e^{(5)} \bm{\varepsilon}_e^{(5)} + \partial \bm{g}_v^T \bm{\pi}_v^{(4)} \bm{\varepsilon}_v^{(4)} + \partial \bm{\eta}^{\prime (4)T}_{x,e} \bm{\pi}_{x,e}^{(4)} \bm{\varepsilon}_{x,e}^{(4)} \\
			\dot{\bm{\mu}}_e^{\prime(4)} &= - \bm{\pi}_{x,e}^{(4)} \bm{\varepsilon}_{x,e}^{(4)}
		\end{split}
	\end{align}
	In short, the 0th-order is subject to: (i) a quantity proportional to the estimated trajectory; (ii) an extrinsic prediction error coming from the elbow, i.e., $\bm{\varepsilon}^{(4)}_e = \bm{\mu}_e^{(4)} - \bm{g}_e(\bm{\mu}_i^{(4)},\bm{\mu}_e^{(3)})$; (iii) a backward extrinsic prediction error coming from the virtual level, i.e., $\bm{\varepsilon}^{(5)}_e = \bm{\mu}_e^{(5)} - \bm{g}_e(\bm{\mu}_i^{(5)},\bm{\mu}_e^{(4)})$; (iv) a visual prediction error, i.e., $\bm{\varepsilon}_v^{(4)} = \bm{y}_v^{(4)} - \bm{\mu}_e^{(4)}$; (v) a backward dynamics error encoding the generated trajectory, i.e., $\bm{\varepsilon}_{x,e}^{(4)} = \bm{\mu}_e^{\prime(4)} - \bm{\eta}_{x,e}^{\prime(4)}$. For a more detailed treatment of inference and dynamics of kinematic configurations in hierarchical settings, see \cite{Priorelli2023b}.
	
	The actions $\bm{a}$ are instead computed by minimizing proprioceptive prediction errors:
	\begin{equation}
		\label{eq:action}
		\dot{\bm{a}} = - \partial_{a} \bm{g}_p^T \bm{\pi}_p \bm{\varepsilon}_p
	\end{equation}
	where $\partial_{a} \bm{g}_p$ performs an inverse dynamics from proprioceptive predictions to actions.
	
	As concerns the discrete model, its hidden states $\bm{s}$ express: (i) whether the agent is at the tool position, at the ball position, or none of the two; (ii) whether the agent has grasped or not the tool. These two factors combine in 6 process states in total. The first factor generates predictions for the extrinsic hidden causes of the hybrid units through likelihood matrices, i.e., $\bm{A}_e^{(4)} \bm{s}$ and $\bm{A}_e^{(5)} \bm{s}$. This allows the agent to synchronize the behavior of both the tool and end effector; additional likelihood matrices can be defined to impose priors for the intrinsic hidden states and at different levels of the hierarchy. The second factor returns a discrete tactile prediction, i.e., $\bm{A}_t \bm{s}$.
	
	Finally, we define a discrete action for each step of the task, and a transition matrix $\bm{B}$ such that the ball can be reached only when the tool has been grasped. Discrete actions are replanned every $10$ continuous time steps, and transitions between discrete states occur dynamically depending on continuous evidence. In the example of Figure \ref{fig:frames}, the transition between reaching the tool and reaching the ball happens after $350$ time steps. The extrinsic hidden causes $\bm{v}_e^{(4)}$ and $\bm{v}_e^{(5)}$ are found by Bayesian model comparison, as explained in Section \ref{section:flexible}:
	\begin{align}
		\label{eq:causes}
		\begin{split}
			\bm{v}_e^{(4)} &= \sigma(\ln \bm{A}_e^{(4)} \bm{s} + \bm{l}_e^{(4)}) \\
			\bm{v}_e^{(5)} &= \sigma(\ln \bm{A}_e^{(5)} \bm{s} + \bm{l}_e^{(5)})
		\end{split}
	\end{align}
	As noted above, $\bm{A}_e^{(4)} \bm{s}$ and $\bm{A}_e^{(5)} \bm{s}$ represent predictions of extrinsic hypotheses made by the discrete model for the end effector and virtual levels, e.g., a higher value of $v_b^{(4)}$ and $v_b^{(5)}$ means that the discrete model wants to reach the ball. Conversely, $\bm{l}_e^{(4)}$ and $\bm{l}_e^{(5)}$ are the bottom-up messages that accumulate continuous log evidence over some time $T$, that is, they provide information whether the end effector is reaching the tool or the ball based on the context. Comparison between such high-level expectations and low-level evidences permits inferring the discrete hidden states based on potential trajectories. In fact, the hidden causes act as additional observations for the discrete model, which infers the states at time $\tau$ by combining them with the tactile observation and the hidden states at time $\tau-1$:
	\begin{equation}
		\bm{s}_{\pi,\tau} = \sigma(\ln \bm{B}_{\pi,\tau-1} \bm{s}_{\tau-1} + \ln \bm{A}_e^{(4) T} \bm{v}_{e,\tau}^{(4)} + \ln \bm{A}_e^{(5) T} \bm{v}_{e,\tau}^{(5)} + \ln \bm{A}_t^T \bm{o}_t)
	\end{equation}
	If we assume for simplicity that the agent's preferences are encoded in a tensor $\bm{C}$ in terms of expected states, the expected free energy breaks down to:
	\begin{equation}
		\label{eq:exp_fe}
		\mathcal{G}_{\pi} \approx \sum_\tau \bm{s}_{\pi,\tau} [\ln \bm{s}_{\pi,\tau} - \ln p(\bm{s}_\tau|\bm{C})]
	\end{equation}
	Computing the softmax of the expected free energy returns the posterior probability over the policies $\bm{\pi}$, which are used to infer the new discrete hidden states at time $\tau+1$.
	
	
	\subsection{\label{analysis}Analysis of model performances}
	
	Figure \ref{fig:frames} illustrates task progress during a sample trial. Although both objects are moving, the discrete model successfully infers and imposes continuous trajectories allowing the agent to operate correctly and achieve its goal. At the beginning of a trial, the beliefs of the hidden states are initialized with the actual starting configuration of the arm. The agent infers two potential kinematic configurations for the tool and the ball. While the two observations of the tool constrain the corresponding inference, the ball belief is only subject to its actual position, thus letting the agent initially overestimate the length of the virtual level. During the first phase, only the tool reaching intention is active: as a consequence, the tool belief constantly biases the arm belief, which in turn pulls the real arm. After 350 steps, both these beliefs are in the same configuration, while the ball belief has inferred the corresponding position. At this point, the tool is grasped, causing the discrete model to predict a different combination of hidden causes. Now, both the tool and arm beliefs are pulled toward the ball belief. After about 800 steps, the agent infers the same configuration for all three beliefs, successfully reaching the ball with the tool's extremity, and tracking it until the trial ends. Note that even during the first reaching movement, the agent continuously updates its configuration in relation to the ball; as a result, the second reaching movement is faster. 
	
	\begin{figure}
		\centering
		\begin{minipage}[b]{.8\textwidth}
			\centering
			{\label{fig:tool_evidence}\includegraphics[width=\textwidth]{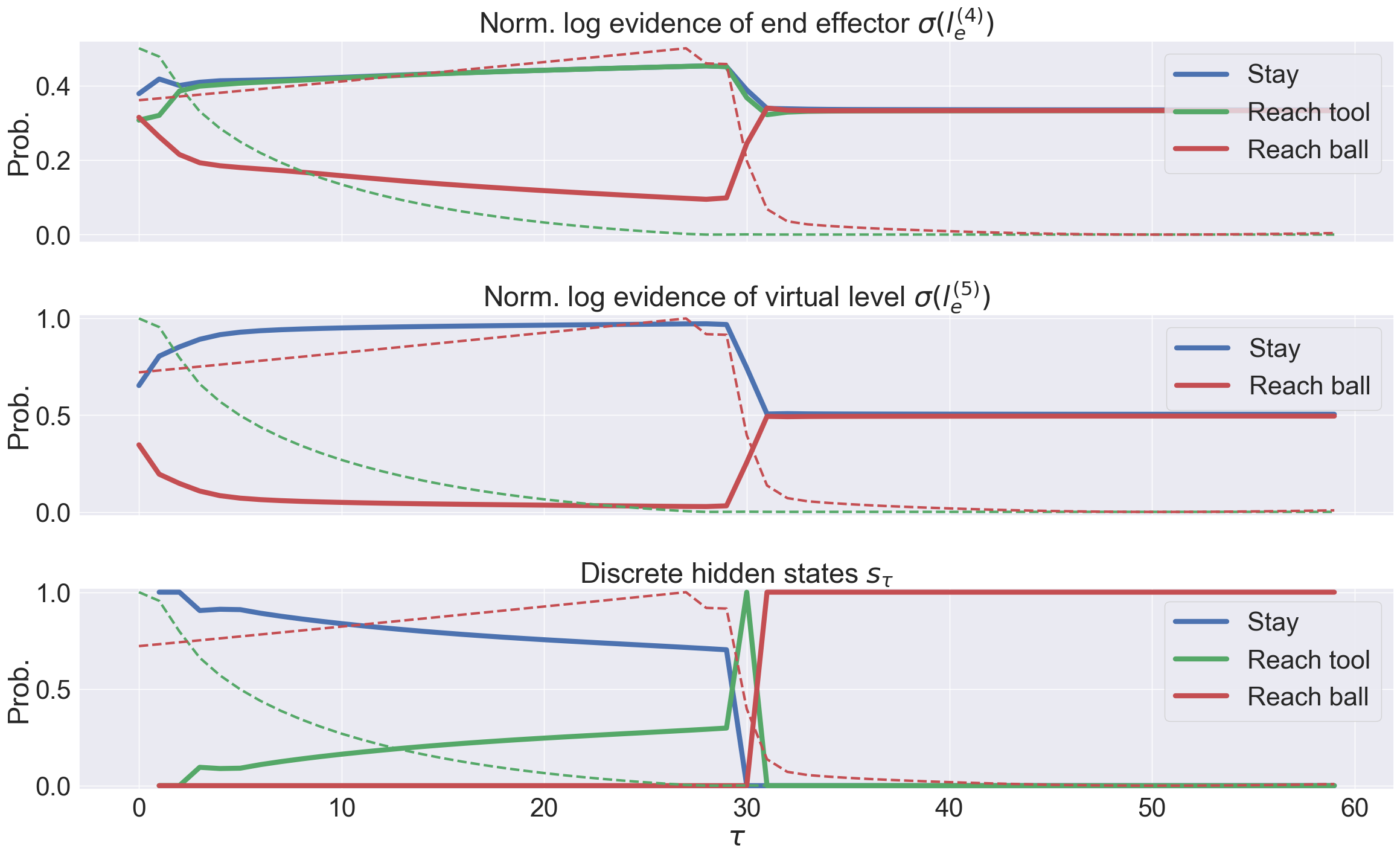}}
		\end{minipage}
		\begin{minipage}[b]{.8\textwidth}
			{\label{fig:tool_dynamics}\includegraphics[width=\textwidth]{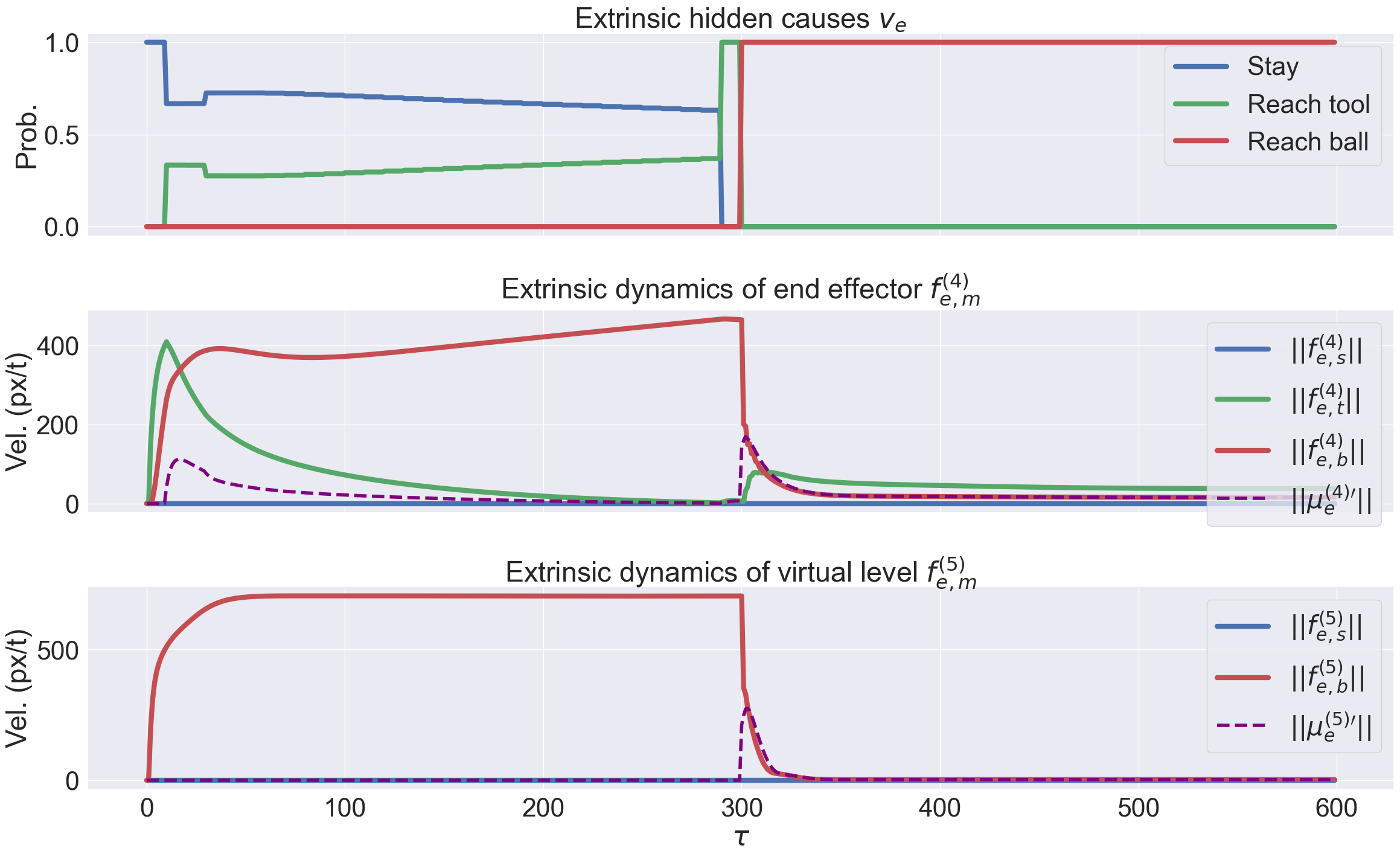}}
		\end{minipage}
		\caption{\textbf{(a)} Normalized log evidences, for $60$ discrete steps $\tau$ (composed, in turn, of 10 continuous time steps), of the end effector's level $\bm{l}_e^{(4)}$ (top), and virtual level $\bm{l}_e^{(5)}$ (middle). Discrete hidden states (bottom). The green and red dashed lines respectively represent the tool's extremity-ball distance, and the tool's origin-end effector distance, normalized to fit in the plots. As explained in Section \ref{appendix:impl}, the agent has two discrete states and two hidden causes related to the steps of the task, i.e., reaching the ball and reaching the tool, with additional \textit{stay} discrete state and cause. \textbf{(b)} Dynamics of extrinsic hidden causes $\bm{v}_e$ (top plot). Norm of extrinsic potential dynamics $||\bm{f}_{e,s}^{(4)}||$ (\textit{stay}), $||\bm{f}_{e,t}^{(4)}||$ (\textit{reach tool}) and $||\bm{f}_{e,b}^{(4)}||$ (\textit{reach ball}), along with estimated dynamics $||\bm{\mu}_{e}^{(4)\prime}||$ for end effector's level (middle plot). Norm of extrinsic potential dynamics $||\bm{f}_{e,s}^{(5)}||$ (\textit{stay}) and $||\bm{f}_{e,b}^{(5)}||$ (\textit{reach ball}), along with estimated dynamics $||\bm{\mu}_{e}^{(5)\prime}||$ for virtual level (bottom plot). The dynamics are plotted for 600 continuous time steps.}
	\end{figure}
	
	The transitions can be better appreciated from Figure \ref{fig:tool_evidence}, showing the bottom-up messages (i.e., accumulated evidences) $\bm{l}_e^{(4)}$ and $\bm{l}_e^{(5)}$ for the last two levels of the hierarchy (i.e., end effector's and virtual levels), and the discrete hidden states $\bm{s}_{\tau}$. As evident, the virtual level does not contribute to the inference of the first reaching trajectory, since this only involves the end effector. Note how the agent is able to dynamically accumulate the evidences over its discrete hypotheses: during the first phase, the evidence $l^{(4)}_{e,t}$ (related to the tool belief at the end effector's level) increases as soon as the end effector approaches the tool's origin, while $l^{(4)}_{e,b}$ and $l^{(5)}_{e,b}$ (related to the ball belief at the end effector's and virtual levels, respectively) decrease as the ball moves away. During the second phase, the latter two rapidly increase as the end effector approaches the ball; finally, every probability of both levels slowly stabilizes as the extrinsic beliefs converge to the same value and the errors are minimized. The slow decrease of the initial state and the fast transition between the two steps are well summarized in the bottom graph. The trajectories of the hidden states show that the agent can plan new trajectories with a high frequency (in this case, $10$ continuous time steps), allowing it to react rapidly to environmental stimuli.
	
	The relationship between hidden causes and continuous dynamics is summarized in Figure \ref{fig:tool_dynamics}, showing the extrinsic potential and estimated dynamics for the end effector's and virtual levels, as well as the dynamics of the extrinsic hidden causes. For simplicity, we considered the same discrete hidden causes for every hierarchical level, and the top plot recapitulates the state of the whole kinematic configuration. Here, we note a similar behavior to the discrete hidden states (i.e., slow decrease of the stay cause and increase of the first reaching movement, and rapid increase of the second reaching movement in the middle of the trial). Note that the agent maintains potential dynamics related to the three intentions for the duration of the whole trial; these dynamics are combined to produce a trajectory that the motor units accomplish. In fact, two spikes are evident in the dynamics of the end effector's level, and one in the dynamics of the virtual level, regarding the ball reaching action.
	
	In order to assess the model performances in dynamic planning tasks, we run three different experiments, each composed of 150 trials. The first experiment assessed the capacity of picking a moving tool and reaching a static target, hence, for each trial we varied the tool velocity. The second experiment assessed the capacity of picking a static tool and tracking a moving target: here, we varied the ball velocity. The third experiment evaluated the performances of the agent in picking a moving tool and tracking a moving target: we hence varied both tool and ball velocities. In all experiments, we randomly sampled tool and ball positions, and their directions where relevant. Also, velocity varied from 0 to 8 pixels per time step. The width and height of the virtual environment was 1300x1300 pixels, twice the total arm length plus the tool length; thus, the ball and tool were out of reach for a significant period of each trial. The duration of each trial was set to 3000 steps. 
	
\begin{figure}
		\centering
		\includegraphics[width=0.85\textwidth]{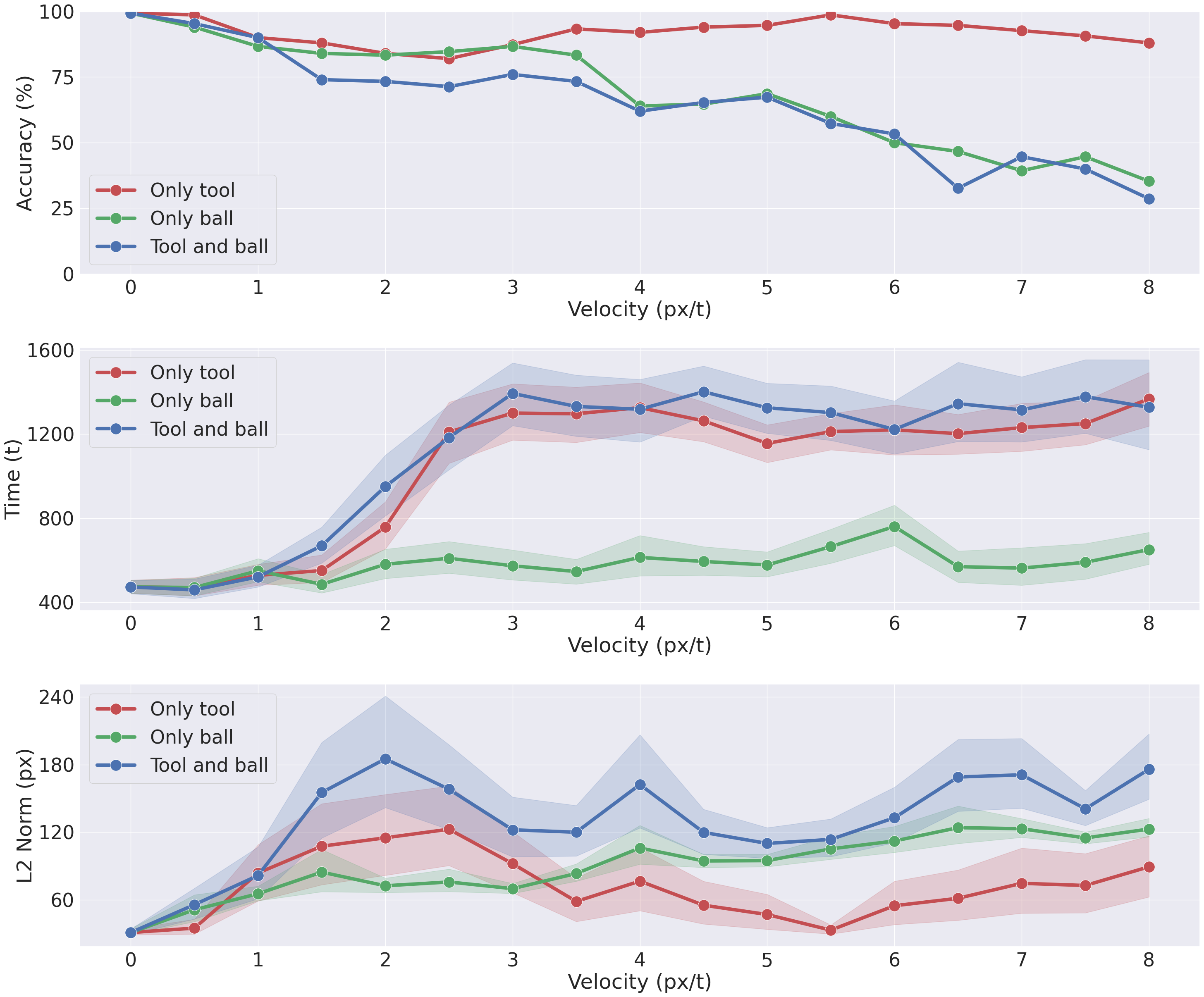}
		\caption{Performances of the deep hybrid model during tool use for three conditions: a moving tool (in red), a moving ball (in green), and moving tool and ball (in blue). \textit{Accuracy} (top), for which we considered a trial successful if the tool was picked and the average ball-tool distance for the last 300 steps was less than 100 pixels. \textit{Time} (middle), measured as the number of steps needed for the tool to be picked and for the ball-tool distance to be less than 100 pixels. \textit{Error} (bottom), measured as the average ball-tool distance for the last 300 steps. For each condition, we aggregated the measures for 150 trials. The middle and bottom plots also show the 95\% confidence interval.}
		\label{fig:plot_scores}
	\end{figure}

	The results of the simulations are visualized in Figure \ref{fig:plot_scores}, showing the task accuracy, the time needed to complete the task, and the average final error (see the caption for more details). With static or slow-varying environments, the agent completes the task in all conditions. The first condition (moving tool) achieved good performances even with high tool velocity, although with low velocities there is a slight decrease in accuracy: since the tool often moves out of reach, the agent cannot accomplish the tool-picking action. This specific behavior is not present in the second condition (moving ball), probably due to the increased operational space which allows the agent to move along the whole environment. However, the performance decreases for high ball velocities also due to the additional difficulty of tracking moving objects. The third combined condition (moving tool and ball) achieved slightly lower performances than the second condition, with a time needed to complete the task similar to the first condition. However, the average tracking error shown in the bottom panel remains restricted even with high ball and tool velocities.
	
\begin{figure}
		\centering
		\includegraphics[width=0.85\textwidth]{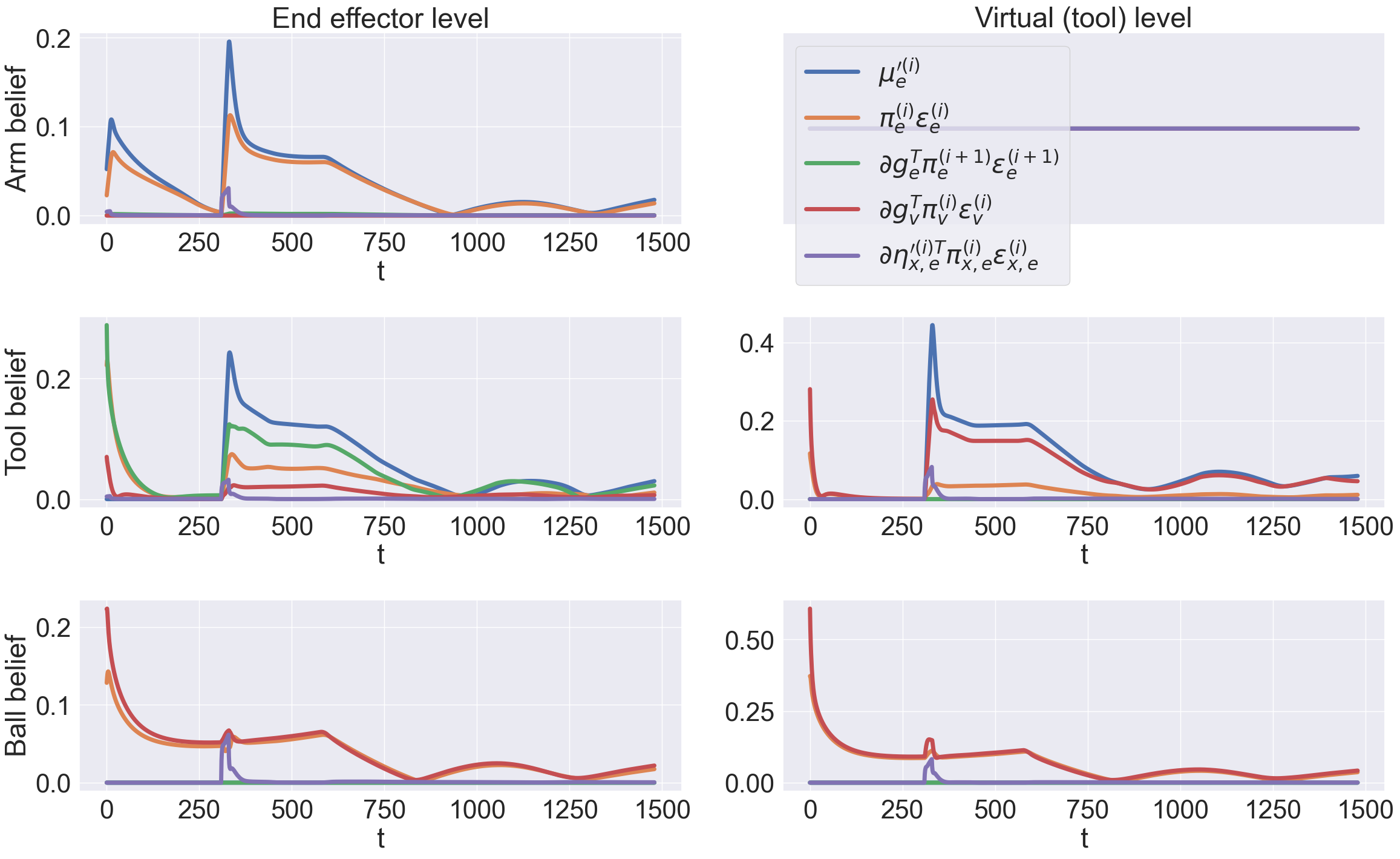}
		\caption{Trajectory of every component of the extrinsic belief updates for the end effector (left panels), and virtual (right panels) levels. Every environmental entity is shown separately. Blue, orange, green, red, and purple lines respectively indicate the 1st-order derivative, the extrinsic prediction error from the previous level, the extrinsic prediction error from the next level, the visual prediction error, and the backward dynamics error.}
		\label{fig:tool_beliefs}
	\end{figure}

	Finally, the dynamic behavior of the extrinsic beliefs can be analyzed from Figure \ref{fig:tool_beliefs}, showing the trajectories, for a sample trial of the third condition, of all the forces that make up the update of Equation \ref{eq:update}, for the last two levels and every environmental entity. The transition between the two phases of the task is here evident: the 1st-order derivative of the arm belief $\bm{\mu}_{e,a}^{\prime (4)}$ (blue line in the top left panel) is non-zero during the whole task, and presents two spikes at the beginning of each phase, signaling an increased prediction error due to the new intention. The arm movement during the first phase is the consequence of the non-zero 1st-order derivative of the tool belief $\bm{\mu}_{e,t}^{\prime (4)}$ (blue line in the middle left panel). The dynamics of the corresponding extrinsic prediction error $\bm{\varepsilon}_{e,t}^{(5)}$ (green line in the middle left panel) combines both this derivative and the visual prediction error of the next level $\bm{\varepsilon}_{v,t}^{(5)}$ (red line in the middle right panel). Note that this extrinsic prediction error does not exist for the arm belief, and that the backward dynamics error $\bm{\varepsilon}_{e,x,a}^{(4)}$ has a smaller impact on the overall update with respect to the 1st-order derivative. The second phase begins with a spike in the 1st-order derivative of the tool belief at the virtual level $\bm{\mu}_{e,t}^{\prime (5)}$ (blue line in the middle right panel), which is propagated back to the previous level as an extrinsic prediction error. Finally, note that the ball belief is only subject to its visual observation and the extrinsic prediction error coming from the previous levels.
	
	\section{Discussion}
	
	We proposed a computational method, based on hybrid active inference, that affords dynamic planning for hierarchical settings. Our goal was twofold. First, to show the effectiveness of casting control problems as inference and, in particular, of expressing entities in relation to a hierarchical configuration of the self. While there could be several ways to combine the units of the proposed architecture, we showed a specific design as a proof-of-concept to solve a typical task: reaching a moving object with a tool. The agent had to rely on three kinds of depth, i.e., it had to dynamically infer its intentions for decision-making, and form different hierarchical generative models depending on the structure and affordances of the entities. The proposed model unifies several characteristics studied in the active inference literature: the modeling of objects, recently done in the context of active object reconstruction \cite{ferraro2022disentangling, vanbergen2022objectbased,VandeMaele2022}; the analyses of affordances in relation to the agent's beliefs \cite{Donnarumma2017}; the modeling of itinerant movements, with a behavior similar to the Lotka-Volterra dynamics implemented in \cite{Friston2011b}; planning and control in extrinsic coordinates \cite{Oliver2021,Friston2010}; inference of discrete states based on continuous signals in dynamic environments, achieved through a different kind of post-hoc Bayesian model selection \cite{Isomura2019} or other various approaches such as bio-inspired SLAM \cite{ccatal2021robot}, dynamic Bayesian networks \cite{Nozari2022}, recurrent switching linear dynamical systems \cite{collis2024learninghybridactiveinference}; the use of tools for solving complex tasks \cite{anil2024}.
	
	Our second goal was to show that a (deep) hierarchical formulation of active inference could lend itself to learning and generalization of novel tasks. Although we used a fixed generative model, we revealed that an advanced behavior is possible by using likelihood and dynamics functions that could be easily implemented with neural connections, and by decomposing the model into small units linked together. In \cite{Priorelli2023f}, a hierarchical kinematic model was used to learn the limbs of an agent's kinematic chain, both during perception and action. The same mechanism could be used to infer the length of tools needed for object manipulation, extending the kinematic chain in a flexible way. Therefore, an encouraging research direction would be to design a deep hybrid model in the wake of PCNs, and let the agent learn appropriate structure and internal attractors for a specific goal via free energy minimization. PCNs have demonstrated robust performance in tasks like classification and regression \cite{Ororbia2022b,salvatori2023braininspired}, while approximating the backpropagation algorithm \cite{Whittington2017, Whittington2019, Millidge2022, salvatori2021predictive}. However, few studies have leveraged the modular and hierarchical nature of predictive coding to model complex dynamics \cite{Stoianov2022,Millidge2023,Jiang2024,nguyen2021,tang2023} or enable interactions with the environment \cite{Millidge2019,Ororbia2022,rao2022active,Fisher2023,Stoianov2018}, mostly done through RL. Well-known issues of deep RL are data efficiency, explainability, and generalization \cite{Millidge2022a}. Instead, the human brain is capable of learning new tasks with a small amount of samples, transferring the knowledge previously acquired in similar situations. Another common criticism is that deep RL lacks explainability, which is of greater concern as AI systems rapidly grow. A viable alternative is to learn a model of the environment \cite{moerland2022modelbased}, e.g., with Bayesian non-parametrics \cite{Stoianov2018}; however, these approaches are still computationally demanding. Albarracin et al. described how active inference may find an answer to the \textit{black box problem} \cite{albarracin2023}, and we further showed how different elements of an active inference agent have practical and interpretable meanings. In this view, optimization of parameters in hybrid models could be an effective alternative to deep RL algorithms, or other approaches in active inference relying on the use of neural networks as generative models.
	
	Besides the fixed generative model, another limitation of the proposed study is that we only used two temporal orders, while a more complete and effective model would make use of a greater set of generalized coordinates \cite{Friston2010gen}. Nonetheless, every aspect we introduced can be extended by considering increasing temporal orders. For instance, discrete variables could depend on the position, velocity, and acceleration of an object, thus inferring a more accurate representation of dynamic trajectories. Also, flexible behavior could be specified in the 2nd temporal order, resulting in a more realistic force-controlled system.
	
	An interesting direction of research regards the generation of states and paths, about which useful indications might come from planning and control with POMDP models \cite{Stoianov2018}. Some implementations of discrete active inference models used additional connections between policies and between discrete hidden states \cite{vandemaele2023objectcentric,FRISTON2024105500}; hence, it might be beneficial to design similar connections in continuous and hybrid contexts as well. In this study, a single high-level discrete model imposed the behavior of every other hybrid unit; an alternative would be to design independent connections between hidden causes such that a high-level decision would be propagated down to lower levels with local message passing. This approach may also provide insights into how, by repetition of the same task, discrete policies adapt to construct composite movements (e.g., a reaching and grasping action) from simpler continuous paths.
	
	
	\section*{Acknowledgments}
	
	This research received funding from the European Union’s Horizon H2020-EIC-FETPROACT-2019 Programme for Research and Innovation under Grant Agreement 951910 to I.P.S. The funders had no role in study design, data collection and analysis, decision to publish, or preparation of the manuscript.
	
	
	\bibliographystyle{unsrt}  
	\bibliography{references}
	
	
	\newpage
	\appendix
	
	
	\clearpage
	\section{\label{appendix:alg}Algorithms}
	
	\begin{algorithm}[h!]
		\caption{Compute expected free energy}
		\label{alg:expected}
		\begin{algorithmic}
			\STATE {\bfseries Input:} \\
			\quad length of policies $N_p$ \\
			\quad discrete hidden states $\bm{s}$ \\
			\quad policies $\bm{\pi}$ \\
			\quad transition matrix $\bm{B}$ \\
			\quad preference $\bm{C}$
			\STATE {\bfseries Output:} \\
			\quad expected free energy $\bm{\mathcal{G}}$
			\STATE
			\STATE $\bm{\mathcal{G}} \gets \bm{0}$
			\FOR{each policy $\bm{\pi}_\pi$}
			\STATE $\bm{s}_{\pi,\tau} \gets \bm{s}$
			\FOR{$\tau=0$ {\bfseries to} $N_p$}
			\STATE $\bm{s}_{\pi,\tau} \gets \bm{B}_{\pi,\tau} \bm{s}_{\pi,\tau}$
			\STATE $\mathcal{G}_\pi \gets \mathcal{G}_\pi + \bm{s}_{\pi,\tau} (\ln \bm{s}_{\pi,\tau} - \ln \bm{C})$
			\ENDFOR
			\ENDFOR
		\end{algorithmic}
	\end{algorithm}
	
	\begin{algorithm}[h]
		\caption{Accumulate log evidence}
		\label{alg:accumulate}
		\begin{algorithmic}
			\STATE {\bfseries Input:} \\
			\quad mean of full prior $\bm{\eta}$ \\
			\quad mean of reduced priors $\bm{\eta}_m$ \\
			\quad mean of full posterior $\bm{\mu}$ \\
			\quad precision of full prior $\bm{\Pi}$ \\
			\quad precision of reduced priors $\bm{\Pi}_m$ \\
			\quad precision of full posterior $\bm{P}$ \\
			\quad log evidences $L_m$ 
			\STATE {\bfseries Output:} \\
			\quad log evidences $L_m$
			\STATE
			\FOR{each reduced model $m$}
			\STATE $\bm{P}_m \gets \bm{P} - \bm{\Pi} + \bm{\Pi}_m$
			\STATE $\bm{\mu}_m \gets \bm{P}_m^{-1} (\bm{P} \bm{\mu} - \bm{\Pi} \bm{\eta} + \bm{\Pi}_m \bm{\eta}_m)$
			\STATE $\mathcal{L}_m \gets \mathcal{L}_m + (\bm{\mu}_m^{T} \bm{P}_{m} \bm{\mu}_m - \bm{\eta}_m^T \bm{\Pi}_{m} \bm{\eta}_m - \bm{\mu}^{T} \bm{P}_x \bm{\mu} + \bm{\eta}^{T} \bm{\Pi} \bm{\eta}) / 2$
			\ENDFOR
		\end{algorithmic}
	\end{algorithm}
	
	\begin{algorithm}[h]
		\caption{Active inference with deep hybrid models}
		\label{alg:active_inference}
		\begin{algorithmic}
			\STATE {\bfseries Input:} \\
			\quad continuous time $T$ \\
			\quad discrete time $\mathcal{T}$ \\
			\quad intrinsic units $\mathcal{U}_i^{(i,j)}$ \\
			\quad extrinsic units $\mathcal{U}_e^{(i,j)}$ \\
			\quad inverse dynamics $\partial_{a} \bm{g}_p$ \\
			\quad proprioceptive precisions $\bm{\Pi}_{y,p}$ \\
			\quad learning rate $\Delta_t$, \\
			\quad action $\bm{a}$
			\STATE
			\FOR{$t=0$ {\bfseries to} $T$}
			\STATE Get observations
			\IF{$t \mod \mathcal{T} = 0$}
			\STATE Update discrete model via Algorithm \ref{alg:discrete_model}
			\ENDIF
			\FOR{each unit $\mathcal{U}_i^{(i,j)}$ and $\mathcal{U}_e^{(i,j)}$}
			\STATE Update intrinsic unit via Algorithm \ref{alg:intrinsic_unit}
			\STATE Update extrinsic unit via Algorithm \ref{alg:extrinsic_unit}
			\ENDFOR
			\STATE Get proprioceptive prediction errors $\bm{\varepsilon}_{y,p}$ from intrinsic units
			\STATE $\dot{\bm{a}} \gets - \partial_{a} \bm{g}_p^T \bm{\Pi}_{y,p} \bm{\varepsilon}_{y,p}$
			\STATE $\bm{a} \gets \bm{a} + \Delta_t \dot{\bm{a}}$
			\STATE Take action $\bm{a}$
			\ENDFOR   
		\end{algorithmic}
	\end{algorithm}
	
	\begin{algorithm}[h]
		\caption{Update discrete model at time $\tau$}
		\label{alg:discrete_model}
		\begin{algorithmic}
			\STATE {\bfseries Input:} \\
			\quad discrete hidden states $\bm{s}$ \\
			\quad policies $\bm{\pi}$ \\
			\quad likelihood matrices $\bm{A}^{(i,j)}$ \\
			\quad transition matrix $\bm{B}$ \\
			\quad prior $\bm{D}$ \\
			\quad accumulated log evidences $\bm{l}^{(i,j)}$
			\STATE {\bfseries Output:} \\
			\quad accumulated log evidences $\bm{l}^{(i,j)}$ \\
			\quad discrete hidden causes $\bm{v}^{(i,j)}$
			\STATE
			\FOR{each unit $\mathcal{U}^{(i,j)}$}
			\STATE $\bm{v}^{(i,j)} \gets \sigma(\ln \bm{A}^{(i,j)} \bm{s} + \bm{l}^{(i,j)})$
			\ENDFOR
			\STATE $\bm{s} \gets \sigma(\ln \bm{D} + \sum_{i,j} \ln \bm{A}^{(i,j) T} \bm{v}^{(i,j)})$
			\STATE Compute expected free energy $\bm{\mathcal{G}}$ via Algorithm \ref{alg:expected}
			\STATE $\bm{\pi} \gets \sigma(-\bm{\mathcal{G}})$
			\STATE $\bm{D} \gets \sum_\pi \pi_{\pi,0} \bm{B}_{\pi,0} \bm{s}$
			\FOR{each unit $\mathcal{U}^{(i,j)}$}
			\STATE $\bm{v}^{(i,j)} \gets \bm{A}^{(i,j)} \bm{D}$
			\STATE $\bm{l}^{(i,j)} \gets \bm{0}$
			\ENDFOR
		\end{algorithmic}
	\end{algorithm}
	
	\begin{algorithm}[h]
		\caption{Update intrinsic unit}
		\label{alg:intrinsic_unit}
		\begin{algorithmic}
			\STATE {\bfseries Input:} \\
			\quad belief of extrinsic hidden states of previous level $\bm{\mu}_{e}^{(i-1)}$ \\
			\quad belief of intrinsic hidden states $\bm{\tilde{\mu}}_{i}^{(i)}$ \\
			\quad intrinsic (discrete) hidden causes $\bm{v}_{i}^{(i)}$ \\
			\quad proprioceptive observation $\bm{y}_p^{(i)}$ \\
			\quad belief of extrinsic hidden states $\bm{\tilde{\mu}}_{e}^{(i)}$\\
			\quad intrinsic dynamics (reduced) functions $\bm{f}_{i,m}^{(i)}$ \\
			\quad proprioceptive likelihood $\bm{g}_p$ \\
			\quad extrinsic likelihood $\bm{g}_e$ \\
			\quad proprioceptive precision $\bm{\Pi}_{y,p}^{(i)}$ \\
			\quad extrinsic precision $\bm{\Pi}_{y,e}^{(i)}$ \\
			\quad intrinsic dynamics precision $\bm{\Pi}_{x,i}^{(i)}$ \\
			\quad learning rate $\Delta_t$
			\STATE {\bfseries Output:} \\
			\quad proprioceptive prediction error $\bm{\varepsilon}_{y,p}^{(i)}$ \\
			\quad extrinsic prediction error $\bm{\varepsilon}_{y,e}^{(i)}$
			\STATE
			\STATE $\bm{\eta}_{x,i}^{(i)\prime} \gets \sum_m v_{i,m}^{(i)} \bm{f}_{i,m}^{(i)}(\bm{\mu}_i^{(i)})$
			\STATE $\bm{\varepsilon}_{y,p}^{(i)} \gets \bm{y}_p^{(i)} - \bm{g}_p(\bm{\mu}_{i}^{(i)})$
			\STATE $\bm{\varepsilon}_{x,i}^{(i)} \gets \bm{\mu}_{x,i}^{(i)\prime} - \bm{\eta}_{x,i}^{(i)\prime}$
			\STATE $\bm{\varepsilon}_{y,e}^{(i)} \gets \bm{\mu}_{e}^{(i)} - \bm{g}_e(\bm{\mu}_{i}^{(i)}, \bm{\mu}_{e}^{(i-1)})$
			\STATE Accumulate log evidence via Algorithm \ref{alg:accumulate}
			\STATE $\dot{\bm{\mu}}_{i}^{(i)} \gets \bm{\mu}_{i}^{(i)\prime} + \partial_{i} \bm{g}_p^T \bm{\Pi}_{y,p}^{(i)} \bm{\varepsilon}_{y,p}^{(i)} + \partial_{i} \bm{g}_e^T \bm{\Pi}_{y,e}^{(i)} \bm{\varepsilon}_{y,e}^{(i)} + \partial_{i} \bm{\eta}_{x,i}^{(i)\prime T} \bm{\Pi}_{x,i}^{(i)} \bm{\varepsilon}_{x,i}^{(i)}$
			\STATE $\dot{\bm{\mu}}_{i}^{(i)\prime} \gets - \bm{\Pi}_{x,i}^{(i)} \bm{\varepsilon}_{x,i}^{(i)}$
			\STATE $\bm{\tilde{\mu}}_{i}^{(i)} \gets \bm{\tilde{\mu}}_{i}^{(i)} + \Delta_t \dot{\bm{\tilde{\mu}}}_{i}^{(i)}$
		\end{algorithmic}
	\end{algorithm}
	
	\begin{algorithm}[h]
		\caption{Update extrinsic unit}
		\label{alg:extrinsic_unit}
		\begin{algorithmic}
			\STATE {\bfseries Input:} \\
			\quad extrinsic prediction error $\bm{\varepsilon}_{y,e}^{(i)}$ \\
			\quad belief of extrinsic hidden states $\bm{\tilde{\mu}}_{e}^{(i)}$ \\
			\quad extrinsic (discrete) hidden causes $\bm{v}_{e}^{(i)}$ \\
			\quad visual observation $\bm{y}_v^{(i)}$ \\
			\quad extrinsic prediction errors of next levels $\bm{\varepsilon}_{y,e}^{(i+1,l)}$ \\
			\quad extrinsic dynamics (reduced) functions $\bm{f}_{e,m}^{(i)}$ \\
			\quad visual likelihood $\bm{g}_v$ \\
			\quad extrinsic precision $\bm{\Pi}_{y,e}^{(i)}$ \\
			\quad extrinsic precisions of next levels $\bm{\Pi}_{y,e}^{(i+1,l)}$ \\
			\quad visual precision $\bm{\Pi}_{y,v}^{(i)}$ \\
			\quad extrinsic dynamics precision $\bm{\Pi}_{x,e}^{(i)}$ \\
			\quad learning rate $\Delta_t$
			\STATE {\bfseries Output:} \\
			\quad belief of extrinsic hidden states $\bm{\tilde{\mu}}_{e}^{(i)}$
			\STATE 
			\STATE $\bm{\eta}_{x,e}^{(i)\prime} \gets \sum_m v_{e,m}^{(i)} \bm{f}_{e,m}^{(i)}(\bm{\mu}_e^{(i)})$
			\STATE $\bm{\varepsilon}_{y,v}^{(i)} \gets \bm{y}_v^{(i)} - \bm{g}_v(\bm{\mu}_{e}^{(i)})$
			\STATE $\bm{\varepsilon}_{x,e}^{(i)} \gets \bm{\mu}_{x,e}^{(i)\prime} - \bm{\eta}_{x,e}^{(i)\prime}$
			\STATE Accumulate log evidence via Algorithm \ref{alg:accumulate}
			\STATE $\dot{\bm{\mu}}_{e}^{(i)} \gets \bm{\mu}_{e}^{(i)\prime} - \bm{\Pi}_{y,e}^{(i)} \bm{\varepsilon}_{y,e}^{(i)} + \sum_l \partial_{e} \bm{g}_e^T \bm{\Pi}_{y,e}^{(i+1,l)} \bm{\varepsilon}_{y,e}^{(i+1,l)} + \partial_{e} \bm{g}_v^T \bm{\Pi}_{y,v}^{(i)} \bm{\varepsilon}_{y,v}^{(i)} + \partial_{e} \bm{\eta}_{x,e}^{(i)\prime T} \bm{\Pi}_{x,e}^{(i)} \bm{\varepsilon}_{x,e}^{(i)}$
			\STATE $\dot{\bm{\mu}}_{e}^{(i)\prime} \gets - \bm{\Pi}_{x,e}^{(i)} \bm{\varepsilon}_{x,e}^{(i)}$
			\STATE $\bm{\tilde{\mu}}_{e}^{(i)} \gets \bm{\tilde{\mu}}_{e}^{(i)} + \Delta_t \dot{\bm{\tilde{\mu}}}_{e}^{(i)}$
		\end{algorithmic}
	\end{algorithm}
	
	
\end{document}